\PassOptionsToPackage{margin=1in}{geometry}
\documentclass[multiauthors,onecolumn]{LaTeXclass/cohere}

\usepackage{ragged2e}
\usepackage{lipsum}
\usepackage{pdfpages}
\usepackage{colortbl}
\usepackage{tabulary}
\usepackage{longtable}
\usepackage{array}
\usepackage{placeins}
\usepackage{tablefootnote}
\usepackage{xcolor}
\usepackage{tcolorbox}
\usepackage{wrapfig}
\usepackage{breqn} % For multiline equations

\usepackage{pifont}
\definecolor{deeppurple}{HTML}{9e02f7}
\definecolor{forestgreen}{HTML}{2e7d43}
\usepackage{multirow}
\usepackage{tabularx}

\usepackage{amsmath,amssymb,amsfonts,mathtools,amsthm}
\usepackage[margin=1in]{geometry}

\usepackage{fancyvrb}
\usepackage{fvextra}
\usepackage{comment}

\usepackage[utf8]{inputenc}

\usepackage{arydshln}

\usepackage{xspace} % useful for proper spacing with variable names (see the \aya definition below)
\usepackage{makecell} % For line breaks in table cells
\usepackage{multirow}

\usepackage{nicematrix}
\usepackage{comment}

\usepackage{bm}
\usepackage{graphicx}
\usepackage{booktabs}
\usepackage{microtype}
\usepackage{hyperref}

\newcommand{\linear}{\textsc{Linear}}
\newcommand{\slerp}{\textsc{Slerp}}
\newcommand{\ties}{\textsc{TIES}}
\newcommand{\simmerge}{\textsc{SimMerge}}

\usepackage{xcolor}
\definecolor{fixedbestblue}{RGB}{30,90,160}
\newcommand{\fixedbest}[1]{\textcolor{fixedbestblue}{#1}}

\setcitestyle{number,square}

\title{SimMerge: Learning to Select Merge Operators from Similarity Signals}

\author{name={Oliver Bolton},affiliation={1}}
\author{name={Aakanksha},affiliation={2}}
\author{name={Arash Ahmadian},affiliation={3}}
\author{name={Sara Hooker},affiliation={4}}
\author{name={Marzieh Fadaee},affiliation={1}}
\author{name={Beyza Ermis},affiliation={1}}
\affiliations{
    \item[1] Cohere Labs
    \item[2] Cohere
    \item[3] Google
    \item[4] Adaption Labs
}

\corresponding[*]{\{\url{beyza}, \url{marzieh}\}\url{{@cohere.com}}}

\abstract{
\justifying

Model merging combines multiple models into a single model with aggregated capabilities, making it a powerful tool for large language model (LLM) development.
However, scaling model merging is challenging: performance depends on the choice of merge operator, model subset, and merge order, often requiring expensive merge-and-evaluate searches.
In this work, we introduce \simmerge{}, a predictive merge-selection method that identifies high-performing merges using inexpensive, task-agnostic similarity signals between models.
Given a small set of unlabeled probes, \simmerge{} extracts functional and structural features to predict the performance of candidate two-way merges, enabling merge operator, order and model subset selection without iterative evaluation.
We show that \simmerge{} consistently outperforms the best fixed merge operator across 7B-parameter LLMs and generalizes to multi-way merges and 111B-parameter LLMs without retraining. We further introduce a bandit variant that supports adding new tasks and operators online. Our results suggest that learning how to merge enables scalable model composition when checkpoint catalogs are large and evaluation budgets are limited.

}

\renewcommand{\FONTauthors}{\bfseries\Large}
\renewcommand{\FONTauthor}{\bfseries\Large}
\begin{document}

\section{Introduction}
\label{introduction}

Model merging is a practical approach for composing checkpoints trained on complementary subsets or domain-specific objectives into a single model that aggregates their capabilities with modest additional data or compute. 
Early weight-space methods such as parameter averaging and related operations improved accuracy and robustness, motivating a broad line of model composition approaches in parameter space \citep{wortsman2022modelsoups,ilharcoediting,matena2022fishermerge}.
More recent work proposes structured or interference-aware rules that make merging more reliable across tasks and initializations \citep{yadav2023ties,huang2024emrmerging,stoica2024zipit}, as well as perspectives that explain or facilitate merging through subspace matching and permutation alignment \citep{tam2023mats,ainsworth2022gitrebasin}.
Tooling further standardizes implementations and recipes, making merging accessible at scale \citep{goddard2024mergekit}.
As modern LLM development produces thousands of fine-tuned checkpoints across tasks, modalities, and languages, model composition is becoming a central requirement for scalable AI systems. In this work, we study merging post-trained checkpoints: fine-tuned LLMs derived from a shared pretrained base.

At this scale, the bottleneck shifts from \textbf{how to merge} to \textbf{how to choose a merge configuration}. From a large catalog, practitioners must select (i) which models to combine, (ii) which operator to apply, and (iii) for multiway merging, in what order. The prevailing approach is empirical search: each candidate requires a merge followed by downstream evaluation, and the cost grows rapidly with the number of models, operators, and merge orders. Moreover, the choice is sensitive where an operator that helps in one domain can harm in another, so a single global recipe is unreliable.

We address this bottleneck with \textbf{predictive merge selection}. We introduce \textbf{\simmerge{}}, a lightweight \emph{trained selector} that predicts the merge operator and, for multiway merges, the merge order using inexpensive pre-merge similarity signals between source checkpoints. Unlike prior work that proposes merge operators or tunes merging hyperparameters \citep{matena2022fishermerge,yadav2023ties,tam2023mats,ainsworth2022gitrebasin,stoica2024zipit,huang2024emrmerging,akiba2025evolutionary}, \simmerge{} learns the selection decision from these signals. Using a small set of unlabeled probes, we compute functional and structural features such as KL divergence between model logits, cosine similarity of weights and attention patterns, and $\ell_2$ distance in parameter space. These signals are far cheaper than running many merge-and-evaluate search yet predictive of downstream outcomes.

\simmerge{} is not a new merge operator; it learns when to apply existing ones. In the pairwise setting, it selects among linear interpolation \citep{wortsman2022modelsoups}, spherical linear interpolation (SLERP) \citep{shoemake1985slerp}, and TIES merging \citep{yadav2023ties}.
The selector is trained offline on previously observed merges to predict which configuration maximizes downstream performance; at deployment it requires only checkpoints and probes, with no downstream evaluation or gradient-based fine-tuning.

\begin{figure*}[t]
  \centering
  \includegraphics[width=0.90\textwidth]{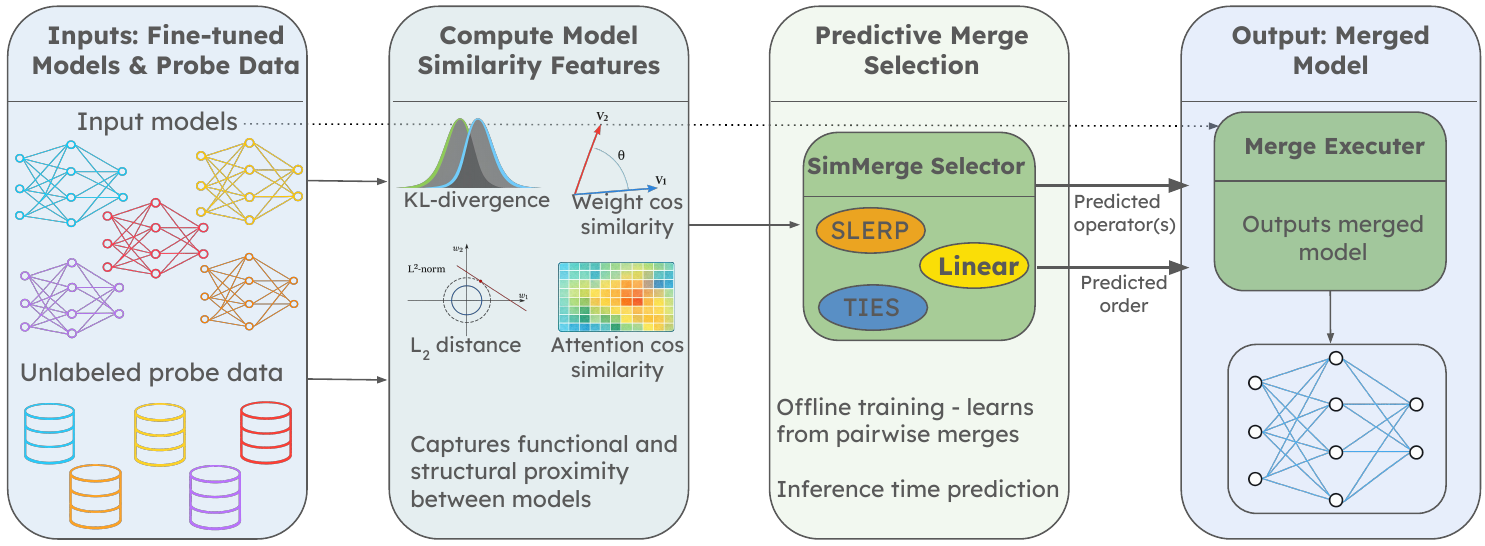}
  \caption{Overview of \simmerge{}. Given a set of domain-specialized checkpoints and small unlabeled probe set for each domain, we compute pre-merge similarity signals, predict the merge operator for each binary merge step and the merge order, and then execute the selected plan once to obtain a single merged model.}
  \label{fig:overview}
\end{figure*}

Predictive selection is especially valuable for multiway merging because common merge operators are not associative: different merge orders can yield different models, and the number of possible orders grows rapidly with the number of checkpoints. Although trained only on pairwise merges, \simmerge{} extends to multiway settings by scoring candidate merge plans, represented as ordered sequences of pairwise merges, using the same similarity features. This enables effective three- and four-way merges without additional supervision.

We consider deployment settings where the model pool and task distribution evolve over time. Offline selectors can drift when new checkpoints or operators appear. We develop an online variant based on a contextual bandit that updates under partial feedback, preserving the low-cost selection step at inference time. Empirically, \simmerge{} is effective on two- to four-way merges of 7B models across multilingual, code, math, and RAG, and transfers to a 111B model without retraining, reducing expert degradation at scale.

Our key contributions are:
\begin{itemize}
\item We introduce \simmerge{}, a \emph{predictive merge selection} method that chooses the model subset, merge operator, and merge order using inexpensive similarity metrics between checkpoints, eliminating costly merge-and-evaluate search.

\item We extend pairwise operator selection to multiway merging by scoring merge plans using the same pairwise similarity features, and show this approach scales from 7B to 111B-parameter models without retraining.

\item We introduce an online contextual bandit variant of \simmerge{} that adapts under partial feedback and supports adding new tasks, models, and operators on-the-fly for evolving checkpoint catalogs.

\item We conduct extensive evaluations across 2-, 3-, and 4-way merges on code, math, multilingual, RAG, and instruction tasks, showing that \simmerge{} consistently outperforms the best fixed merge operator. Averaged over the tasks, \simmerge{} closes $65.0\%$ of the expert-off-domain performance gap, compared to $41.8\%$ for the best single fixed operator.
\end{itemize}

\section{Methodology}\label{sec:method}

We consider a catalog of post-trained checkpoints $\mathcal{M}=\{m_1,\dots,m_n\}$ derived from a shared pretrained base, and a set of tasks $\mathcal{T}=\{t_1,\dots,t_{|\mathcal{T}|}\}$.
Evaluating a model $m\in\mathcal{M}$ on a task $t\in\mathcal{T}$ yields a scalar \emph{utility} $U(m,t)$ computed from the task's evaluation metric on held-out data.
Our goal is to construct a single composite model $\tilde m$ by merging checkpoints from $\mathcal{M}$ so that $U(\tilde m,t)$ is high for a target task $t$, or high on average across a set of tasks.

\subsection{Merge Operators and Plans}
\label{subsec:merge-plans}

\textbf{Binary merge operators.}
Pairwise merges are performed using binary merge operators $o\in\mathcal{O}$. Given two checkpoints $(m_a,m_b)$ and a mixing coefficient $\alpha\in[0,1]$, the merged checkpoint is denoted by $o(m_a,m_b;\alpha)$.
In all experiments we use equal-weight merges with $\alpha=0.5$ and the operator set
$\mathcal{O}=\{\linear{},\slerp{},\ties{}\}$. Formal operator definitions are given in Appendix~\ref{app:merge-operators}.

\textbf{Merge plans.}
A \emph{merge plan} specifies an ordered sequence of checkpoints and the merge operator used at each binary step.
Given an ordered plan $\pi = (m_{i_1}, m_{i_2}, \dots, m_{i_k})$, the merged model is obtained by sequentially applying binary merge operators: 
\[
\begin{cases}
M_1 = m_{i_1},\\
M_j = o_j(M_{j-1}, m_{i_j}; \alpha), & j = 2, \dots, k,
\end{cases}
\]
where $o_j\in\mathcal{O}$ and the final merged model is $\tilde m = M_k$.

Many merge operators are not associative, including \slerp{} and \ties{}, so both the order in $\pi$ and the per-step operator choices $\{o_j\}$ can affect the resulting model and its utility $U(\tilde m,t)$. The number of candidate plans grows rapidly with the number of checkpoints: for a fixed subset of $k$ models there are $k!$ possible orders and $|\mathcal{O}|^{k-1}$ operator sequences, making merge-and-evaluate search infeasible at scale. \simmerge{} addresses this bottleneck by selecting operators and merge plans from pre-merge similarity signals.

\subsection{Similarity Features}
\label{subsec:similarity-features}

For each task $t\in\mathcal{T}$ we draw a small unlabeled probe set $\mathcal{P}_t$ from the input distribution of $t$. For every \emph{ordered} pair of checkpoints $(m_a,m_b)$ and task $t$, \simmerge{} constructs a feature vector $x(m_a,m_b,t)\in\mathbb{R}^m$ using $\mathcal{P}_t$ and the checkpoint weights.
These metrics are inexpensive to compute and capture both functional behavior and structural alignment. (Probe sizes are reported in Section~\ref{sec:exp-setup}.)

%We use ordered pairs because some metrics are asymmetric, so $x(m_a,m_b,t)\neq x(m_b,m_a,t)$ in general. 
We first compute functional similarity signals on $\mathcal{P}_t$, including the KL divergence between predictive distributions, $D_{\mathrm{KL}}(p_a\,\|\,p_b)$, and cosine similarity between intermediate activations $h_a^{(\ell)}$ and $h_b^{(\ell)}$ at each layer $\ell$, averaged over inputs and layers.
We then compute weight-based measures that compare checkpoints directly in parameter space, including cosine similarity between flattened parameter vectors, Euclidean distance between parameters, and parameter norms. We additionally include cosine similarity of attention patterns as a separate feature channel.

Many metrics yield either a scalar or a short sequence over layers or modules. To obtain a fixed-dimensional representation, we summarize sequence-valued metrics using mean, median, and selected
quantiles, and concatenate all summaries into $x(m_a,m_b,t)$.
By default, we append a task encoding $c(t)\in\mathbb{R}^{d_c}$ and use
\[
\tilde{x}(m_a,m_b,t)=x(m_a,m_b,t)\oplus c(t)\in\mathbb{R}^{m+d_c}
\]
as input to all learned components.
We also evaluate a task-agnostic variant that omits $c(t)$ and report the comparison in Appendix~\ref{app:classifier-acc}.
The improvement from task encoding is modest but consistent, so we use it by default. Full details of the similarity metrics and aggregation procedures are provided in Appendix~\ref{app:similarity-metrics}.

\textbf{Computation cost.} Probe-based metrics require only forward passes over small probe sets. We cache per-checkpoint probe outputs so pairwise comparisons reduce to inexpensive post-processing.
Weight-based metrics require a single pass over parameters, and the overall cost is far lower than running merge-and-evaluate searches.

\subsection{Predictive Merge Selection}
\label{subsec:predictive-selection}

Using the precomputed similarity features, \simmerge{} predicts merge operators and scores candidate merge plans without executing merges during selection. Throughout this section we fix the mixing coefficient to $\alpha=0.5$ and omit it from the notation, writing $o(m_a,m_b)$ for $o(m_a,m_b;0.5)$.

\subsubsection{Pairwise utility prediction.}
\simmerge{} is built around a learned utility predictor that estimates the performance of a merge configuration for a given task. In the pairwise setting, for an ordered pair of checkpoints $(m_a,m_b)$ and task $t\in\mathcal{T}$, we define a utility predictor
\[
f_{\mathrm{plan}}:\mathbb{R}^{m+d_c}\rightarrow \mathbb{R}^{|\mathcal{O}|},
\]
which predicts the utility of merging $m_a$ and $m_b$ under each merge operator $o\in\mathcal{O}$. The predictor takes as input pre-merge similarity features $\tilde{x}(m_a,m_b,t)$ and is trained on offline two-way merges, where each operator is applied and the downstream utility $U(o(m_a,m_b),t)$ is observed.
Training minimizes a regression loss toward the observed utilities for all operators:
\[
\widehat{U}(o(m_a,m_b),t)=f_{\mathrm{plan}}(\tilde{x}(m_a,m_b,t))_o.
\]
At inference time, we select the operator with the highest predicted utility,
\[
\hat{o}=\arg\max_{o\in\mathcal{O}} \widehat{U}(o(m_a,m_b),t),
\]
replacing exhaustive merge-and-evaluate search.

\subsubsection{Multi-way merge plan scoring.} 
Multi-way merging generalizes this formulation by selecting an entire merge plan that maximizes predicted utility.
A $k$-way merge plan $\pi = (m_{i_1}, \ldots, m_{i_k})$ is represented as an ordered sequence of binary merge steps.
\simmerge{} estimates the utility of a plan by recursively predicting the utility of each intermediate merge, using the same plan scorer to select operators at each step.

Because intermediate merged models are not explicitly constructed during scoring, similarity features involving an intermediate model $M_{1:k-1}$ are approximated using features propagated from the original pairwise similarity table (see Appendix~\ref{app:metric-propagation}).
Different merge orders induce different sequences of predicted utilities, allowing \simmerge{} to capture the effects of both operator choice and merge order.

At test time, \simmerge{} enumerates or samples a small set of candidate merge plans, estimates their predicted utilities, and selects the plan with the highest predicted utility:
\[
\hat{\pi} = \arg\max_{\pi} \widehat{U}(\pi, t).
\]
The selected plan is then executed as a sequence of binary merges with the step-wise operators $\{\hat{o}_j\}$ chosen by the pairwise predictor.

\subsection{Bandit View and Evaluation Protocol}
\label{sec:bandit-view}

The offline selector performs well on the distribution of model pairs and tasks seen during training, but it must be retrained when new checkpoints, tasks, or operators are introduced.
To support such scenarios without retraining the similarity feature computation, we cast operator selection as a contextual bandit and add a neural-linear bandit layer on top of the precomputed similarity features. As new model pairs and tasks appear, we compute their contexts from the same fixed feature pipeline and update the bandit online using the newly observed rewards. 

\textbf{Contextual bandit formulation.}
Each merge step, either a standalone two-way merge or a step within a multiway plan, defines a decision round. We observe a context vector $s\in\mathbb{R}^{m+d_c}$ derived from pre-merge similarity features. For pairwise merges, $s=\tilde{x}(m_a,m_b,t)$.
For multiway merges, $s$ is constructed analogously and includes propagated similarity features for intermediate merges. We then choose an action $a$ from the operator set $\mathcal{O}$, apply the corresponding merge operator, and observe a scalar reward $r(a)$ given by downstream evaluation.
Rewards for unchosen operators are not observed. The objective is to learn a policy $\pi(s)$ that maximizes cumulative reward, or equivalently minimizing regret relative to an oracle that always selects the best operator for each context.

We adopt a neural-linear design. An MLP feature map $g_\phi$ transforms the context $s$ into a representation $z(s)=g_\phi(s)$. We warm-start $g_\phi$ using the logged pairwise data described below, then keep it fixed during online adaptation. On top of $z(s)$ we fit a linear contextual bandit: for each operator $a \in \mathcal{O}$ we assume a linear reward model
\[
\mathbb{E}[r(a) \mid z(s)] \approx w_a^\top z(s),
\]
with an unknown parameter vector $w_a$ and a Gaussian posterior 
$\mathcal{N}(\hat{w}_a,\Sigma_a)$ maintained via Bayesian linear regression.
We consider both LinUCB and linear Thompson sampling (LinTS)~\citep{abbasi2011improved,agrawal2013thompson}. Empirically, LinTS produces lower regret and higher downstream performance in our setting, and we therefore use it as our main bandit variant.
Since we have a small number of operators and a low-dimensional representation, posterior updates can be done with rank-one updates in $O(d^2)$ time per round, where $d$ is the dimension of $z(s)$.

\textbf{Warm-start and online adaptation.}
The bandit is initialized using the fully-observed pairwise merge dataset from the offline setting. For each historical 2-way merge we observe the context $s$ and the utilities $\$U(\mathsf{a}(m_a, m_b),t)$ for all operators $a \in \mathcal{O}$, which corresponds to full-information feedback.
This warm-start phase anchors the reward model before any online interaction and reduces the amount of exploration required when new tasks or models are introduced.

After warm-start, we introduce a distribution shift by adding a checkpoint trained on a different task that did not appear in the logged data. 
For each new merge involving this checkpoint, we compute the context $s$ from pre-merge similarity signals, query the bandit to select an operator $\mathsf{a}$, execute the merge, and observe the resulting reward $r(a)$. Only the posterior corresponding to the chosen arm $a$ is then updated. This is the partial-feedback regime where counterfactual rewards for unchosen operators are not observed.

\section{Experimental Setup}
\label{sec:experiments}

We evaluate \simmerge{} on four domains: code generation, mathematical reasoning, multilingual understanding and RAG.
All experts share a common pretrained base (Command-A 7B or 111B~\citep{cohere2025command}), ensuring differences arise from fine-tuning and merging rather than capacity.

For a target task $t$ and a set of $k$ models, we compare three fixed merge operators -- \linear, \slerp, and \ties{} against our learned selector, \simmerge{}. Fixed operators apply the same rule at every merge step, using a predetermined order when $k>2$. 
In contrast, \simmerge{} chooses an operator (and, for $k>2$, a merge order) based on pre-merge similarity features, allowing performance gains without exhaustive merge-and-evaluate search.

\subsection{Experts and Auxiliary Models}
\label{sec:exp-experts}

For each domain $t$, we fine-tune a shared base model on task-specific data and designate the best resulting checkpoint as the \emph{task expert} $m^{\text{exp}}_t$.
When evaluating task $t$, any model not fine-tuned on $t$ is treated as an \emph{auxiliary} model; thus, experts from other domains act as auxiliaries for $t$.
In all merge configurations, we combine one task expert with one or more auxiliary models.
We report the standalone performance of task experts and auxiliary models as reference upper/lower bound baselines for task $t$, noting the best-performing expert or auxiliary may vary across metrics or configurations.

Across offline experiments at 7B, we use 85 domain-specialized checkpoints: 23 code, 24 math, 24 multilingual, and 15 RAG. For online bandit experiments, we add 15 instruction-tuned 7B checkpoints, yielding 100 distinct 7B models in total. At the 111B, we evaluate on 18 additional task-specific checkpoints (5 code, 5 math, 4 multilingual, and 4 RAG).

\subsection{Merge Configurations}\label{sec:exp-merges}

\textbf{Pairwise merges.}
We first study pairwise merges between a task expert and a single auxiliary. These experiments provide training data for our selectors and test whether similarity signals are sufficient to predict which operator works best for a given expert-auxiliary pair.

The pairwise dataset used for offline training comprises $240$ distinct expert-auxiliary merges.
Among these training merges, the best-performing operator is Linear in $96$ cases, SLERP in $88$, and TIES in $56$, indicating that no single fixed operator dominates.
For offline evaluation, we construct a held-out test set of $60$ additional expert-auxiliary pairs.
For each of these test pair we evaluate all three operators and log the best-performing method. We use this label to assess the selector's predictive accuracy, with detailed classifier results reported in Appendix~\ref{app:classifier-acc}.
All evaluation splits are disjoint from the pairwise training data at the pair level where no expert-auxiliary pair used to fit the selector appears in the held-out pairwise test set or in any 3-way or 4-way configuration.

\textbf{Multi-way merges.}
We then consider 3-way and 4-way merges, selecting models so that there is at most one expert for each task within a single merge. We merge one task expert with auxiliaries from different domains and apply the same set of operators: Linear, SLERP, TIES, \simmerge{}.
For $k>2$ the merge order affects the result, we therefore evaluate both random-order baselines and the order proposed by our multiway selector.
In total we construct $145$ distinct 3-way merge configurations and $130$ distinct 4-way configurations.
Each configuration is evaluated with all three fixed operators and with \simmerge{}, yielding a large corpus of merge outcomes on which to assess the quality of multiway planning.

\textbf{Scaling to a larger model.}
To test transfer across scales, we repeat the 3-way merging experiments with a substantially larger base model with 111B parameters, using the same similarity features and selector architectures trained on the smaller 7B models. At 111B, we construct an evaluation set of $100$ 3-way merge configurations to test whether a selector trained at one scale can be reused at a larger scale without retraining.

\textbf{Online evaluation merges.}
For the linear bandit experiments, we generate a separate test set of merge configurations that includes the additional \emph{instruct} checkpoints.
This bandit evaluation set contains $60$ merge instances for 2-, 3-, and 4-way merges and is constructed to be disjoint from the pairwise training data at the pair level, meaning no ordered model pair appearing in the bandit evaluation appears in the offline pairwise training set.

\subsection{Evaluations}
\label{sec:exp-evals}

We evaluate expert and merged models on held-out evaluation sets covering our four core domains, plus an instruction-following suite used in the bandit experiments.

For each task $t$, we evaluate on a small collection of standard benchmarks: .
Math is evaluated on \textsc{MATH}~\citep{hendrycks2021measuring} and \textsc{GSM8K}~\citep{zeng2023mr}; code generation on \textsc{HumanEval}~\citep{chen2021evaluating} and \textsc{MBPP+}~\citep{liu2023your}; multilingual understanding on \textsc{MGSM}~\citep{shi2022language} and an internal multilingual QA suite; retrieval-augmented generation (RAG) on \textsc{TauBench}~\citep{yao2025taubench} and \textsc{BFCL}~\citep{patil2025bfcl}; and instruction-following on \textsc{IFEval}~\citep{zhou2023instruction} (details in Appendix~\ref{app:evals}). Metrics follow standard task conventions; we repeat each evaluation three times with different seeds and report the mean. For each model and task $t$, we report a single task score given by the unweighted mean across $t$'s benchmarks.

\subsection{Metrics}
\label{sec:exp-metrics}

We evaluate merge quality using both absolute task performance and normalized percentage change relative to two natural baselines: (i) the task expert and (ii) the auxiliary model(s) involved in a merge. Reporting normalized change relative to these baselines enables fair comparison across tasks with different difficulty and score scales.

For a task $t$, we define the \emph{expert baseline} score as
\[
s_{\text{expert}}(t) := \mathrm{score}(\text{expert}_t, t),
\]
where $\text{expert}_t$ is the model fine-tuned specifically for task $t$.

We define the \emph{auxiliary baseline} score $s_{\text{aux}}(t)$ differently depending on the merge configuration.
For pairwise merges,
\[
s_{\text{aux}}(t) := \mathrm{score}(\text{aux}_t, t),
\]
where $\text{aux}_t$ is the single auxiliary model paired with the expert.
For multiway merges, we define
\[
s_{\text{aux}}(t) := \frac{1}{|\mathcal{A}_t|} \sum_{m' \in \mathcal{A}_t} \mathrm{score}(m', t),
\]
where $\mathcal{A}_t$ is the set of auxiliary models included in the merge configuration.

For a merged model $m$ evaluated on task $t$, we compute normalized percentage change relative to each baseline:
\begin{align*}
\Delta_{\text{expert}}(m,t)
&=
100 \cdot \frac{\mathrm{score}(m,t) - s_{\text{expert}}(t)}
{s_{\text{expert}}(t)}, \\
\Delta_{\text{aux}}(m,t)
&=
100 \cdot \frac{\mathrm{score}(m,t) - s_{\text{aux}}(t)}
{s_{\text{aux}}(t)}.
\end{align*}

These metrics quantify relative gains over the expert and auxiliary baselines, respectively, and account for differences in scale across tasks.

In addition, we report the \emph{gap closed} metric, which measures how much of the performance gap between the auxiliary baseline and the task expert is recovered by the merged model:
\[
\mathrm{GapClosed}(m, t)
=
100 \times
\frac{\mathrm{score}(m,t) - s_{\text{aux}}(t)}
{s_{\text{expert}}(t) - s_{\text{aux}}(t)}.
\]
Under this normalization, $0\%$ corresponds to auxiliary performance and $100\%$ corresponds to expert performance. Values above $100\%$ indicate that the merged model exceeds the expert, while negative values indicate performance below the auxiliary baseline.

All normalized metrics are aggregated across tasks by macro-averaging over $t \in \mathcal{T}$. Absolute performance is also reported by macro-averaging the raw task scores to anchor normalized percentage change to the original evaluation scales.

\subsection{Selector Architecture and Training Details}
\label{sec:exp-setup}

\textbf{Offline classifier.}
We train a lightweight MLP that takes the similarity features $\tilde{x}(m_a,m_b,t)$ as input and outputs the predicted utility for the merge on task $t$. We use a two-layer network with ReLU activations and Adam optimization. We tune basic hyperparameters (hidden width, learning rate, batch size, dropout) on a held-out validation subset of the pairwise merge dataset and use early stopping on validation accuracy.

\textbf{Bandit-based selector.}
%For the online experiments 
We adopt the neural-linear contextual bandit described in Section~\ref{sec:bandit-view}.
We instantiate the bandit feature map as a separate MLP encoder $g_\phi$, and take $z(\tilde{x})$ to be its final hidden-layer representation.
We warm-start this encoder and the per-operator Bayesian linear models using the training pairwise merge data described in Section~\ref{sec:bandit-view}, and then keep the encoder fixed during online adaptation.
On top of $z(\tilde{x})$, we maintain a Bayesian linear model for each operator and use linear Thompson sampling~\citep{abbasi2011improved,agrawal2013thompson} as our primary bandit policy, tuning its exploration and regularization hyperparameters on a validation split of the logged data.
This policy is then used to select operators when new tasks or models, for instance, a new set of instruct checkpoints are introduced, without retraining the encoder.

%%========================================================================================================

\section{Results}\label{sec:results}

Across code, math, multilingual, and RAG, predictive merge selection with \simmerge{} consistently improves the performance of the final merged model.
We show these gains for pairwise and multiway merges in Sections~\ref{sec:results-pairwise} and~\ref{sec:results-multiway}, respectively. In Section~\ref{sec:results-order}, we demonstrate the importance of merge-order prediction for \simmerge{}. Section~\ref{sec:results-111b} shows that a \simmerge{} selector trained on 7B pairwise merges transfers to 111B 3-way merges without retraining. Finally, Section~\ref{sec:bandit-results} considers an online setting with distribution shift and shows that a bandit variant learns to select operators efficiently, closely approaching oracle selection.

\subsection{Learning to Choose the Right Merge Operator in the Pairwise Setting}
\label{sec:results-pairwise}

We begin by examining the central question underlying merge selection: \textit{given two fine-tuned models, can we predict which operator will produce the best merged model?}
Pairwise (2-way) merges offer the cleanest setting to answer this question and form the basis for training and validating \simmerge{}.

For each expert-auxiliary checkpoint pair, we compute pre-merge similarity features and use \simmerge{} to select among \linear{}, \slerp{}, and \ties{}. We then compare the resulting merged model against each fixed operator. This design isolates the contribution of operator selection without changing other merging hyperparameters.

%% FIGURE 2%%
\begin{figure}[h!]
  \centering
  \includegraphics[width=0.90\linewidth]{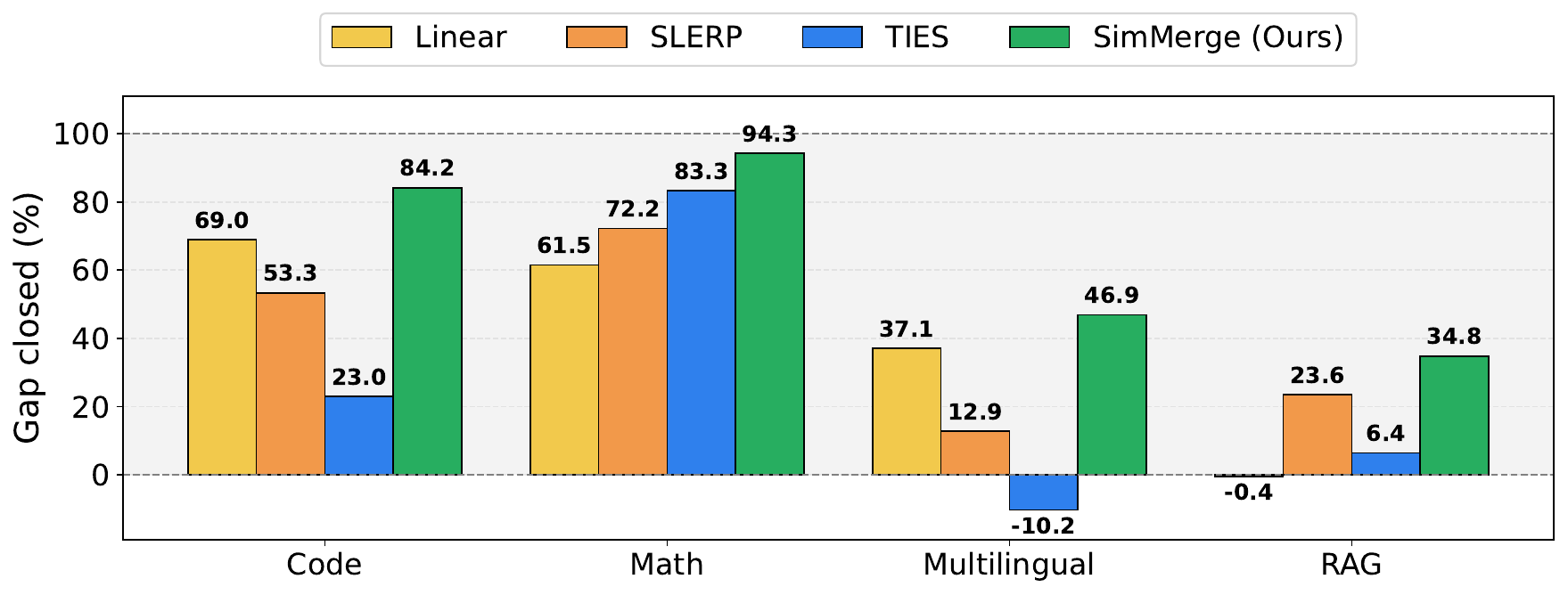}
  \caption{Percentage of the expert–auxiliary performance gap closed by each merge method across Code, Math, Multilingual, and RAG tasks. SimMerge consistently recovers a larger fraction of expert performance than fixed merge operators across all domains.}
  \label{fig:pairwise-gap-closed}
\end{figure}
%%%%%%%%%%%%%

Figure~\ref{fig:pairwise-gap-closed} reports the fraction of the expert-auxiliary performance gap closed by each merge method across domains. Fixed operators exhibit substantial task dependence, with different operators excelling on different tasks and no single operator performing well across all domains. For example, \linear{} is competitive on Code (69.0\%) but slightly underperforms the auxiliary baseline on RAG (-0.4\%), while \ties{} is strong on Math (83.3\%) yet regresses below the auxiliary baseline on Multilingual (-10.2\%).

%% FIGURE 3%%
\begin{figure}[ht!]
    \begin{minipage}[t]{1.0\linewidth}
    \hspace{-3mm}
    \centering
    \includegraphics[width=0.90\linewidth]{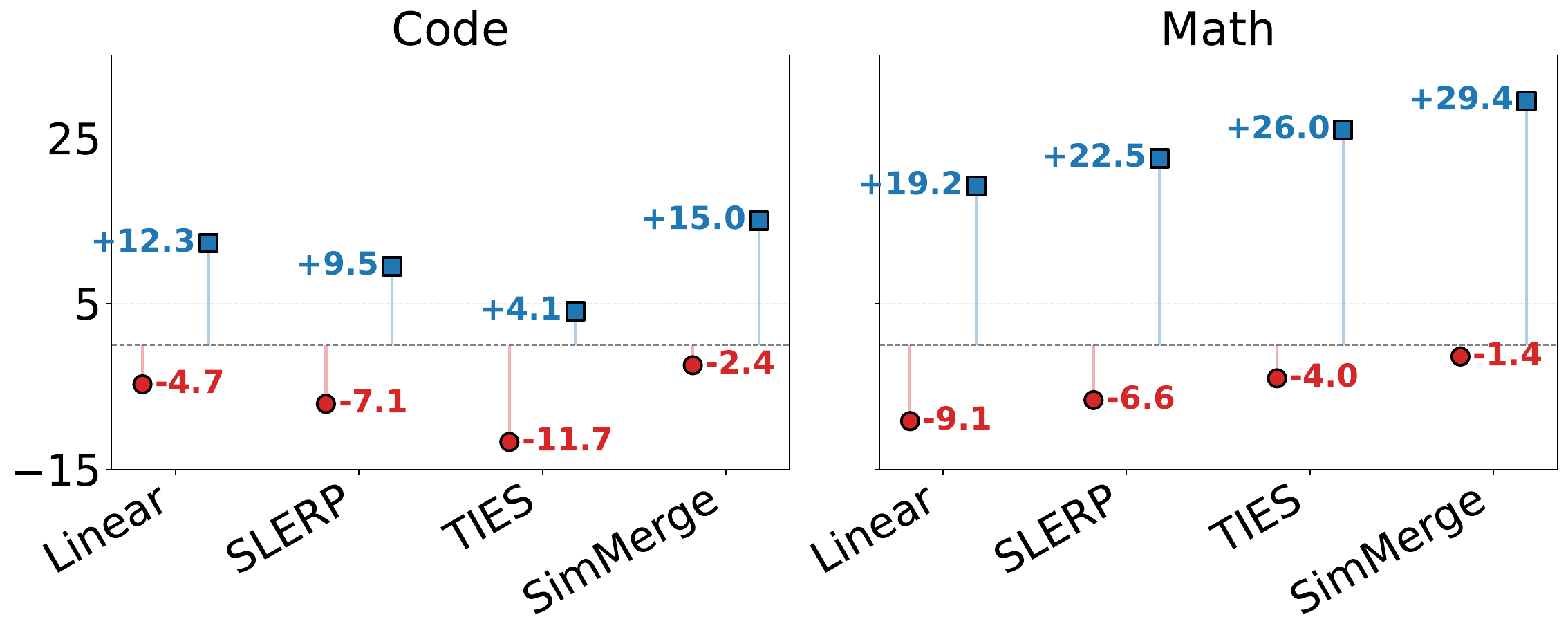}
    \end{minipage}
    \begin{minipage}[t]{1.0\linewidth}
    \hspace{-3mm}
    \centering
    \includegraphics[width=0.90\linewidth]{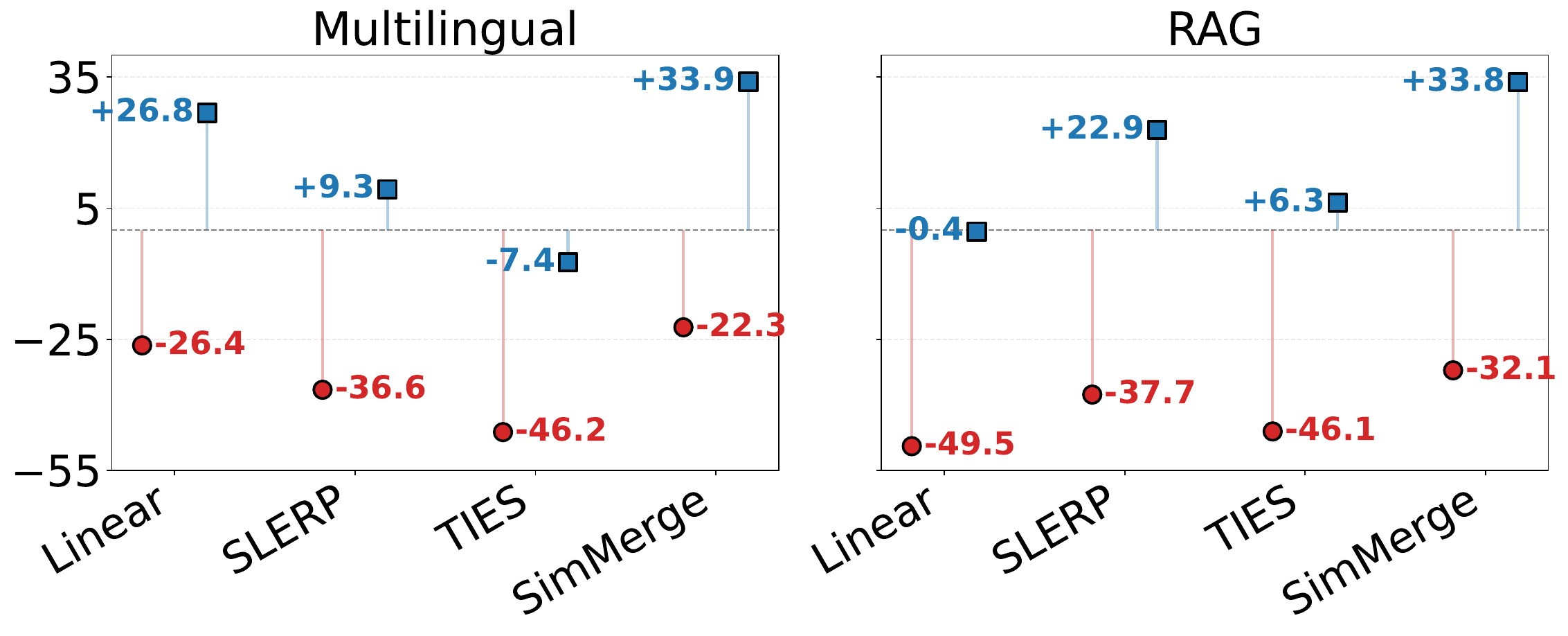}
    \end{minipage}
  \caption{Per-task normalized percentage change in performance for each merge method. Blue markers show $\Delta_{\text{aux}}$ and red markers show $\Delta_{\text{expert}}$ (closer to $0$ indicates less degradation).}
  \label{fig:pairwise-percent}
\end{figure}
%%%%%%%%%%%%%

In contrast, \simmerge{} consistently closes the largest fraction of the expert gap in all four domains. It achieves 84.2\% \emph{GapClosed} on Code, 94.3\% on Math, 46.9\% on Multilingual, and 34.8\% on RAG, outperforming the strongest fixed operator in each case. Macro-averaged across the four domains, \simmerge{} closes 65.0\% of the expert-auxiliary gap, compared to 41.8\% for the best single fixed operator (\linear).

The advantages of predictive operator selection is further illustrated in Figure~\ref{fig:pairwise-percent}, which shows the normalized percentage change in merged model performance relative to the auxiliary baseline (blue) and the task expert (red). Across all four domains, \simmerge{} achieves the largest gains over the auxiliary model while incurring the smallest degradation relative to the expert.

These pairwise results indicate that similarity-driven operator consistently chooses an effective merge rule for each expert-auxiliary pair, producing merged models that improve over the auxiliary baseline while remaining closest to expert performance across all domains.
Appendix~\ref{sec:appendix-tails} further analyzes how similarity signals correlate with merge outcomes and highlights the performance tails where fixed operators exhibit large regressions, underscoring why per-instance operator selection is critical for robust merging.

\subsection{From Pairwise Selection to Multi-Way Merges}
\label{sec:results-multiway}

\begin{figure}[h!]
  \centering
  \hspace{-3mm}
  \includegraphics[width=0.90\linewidth]{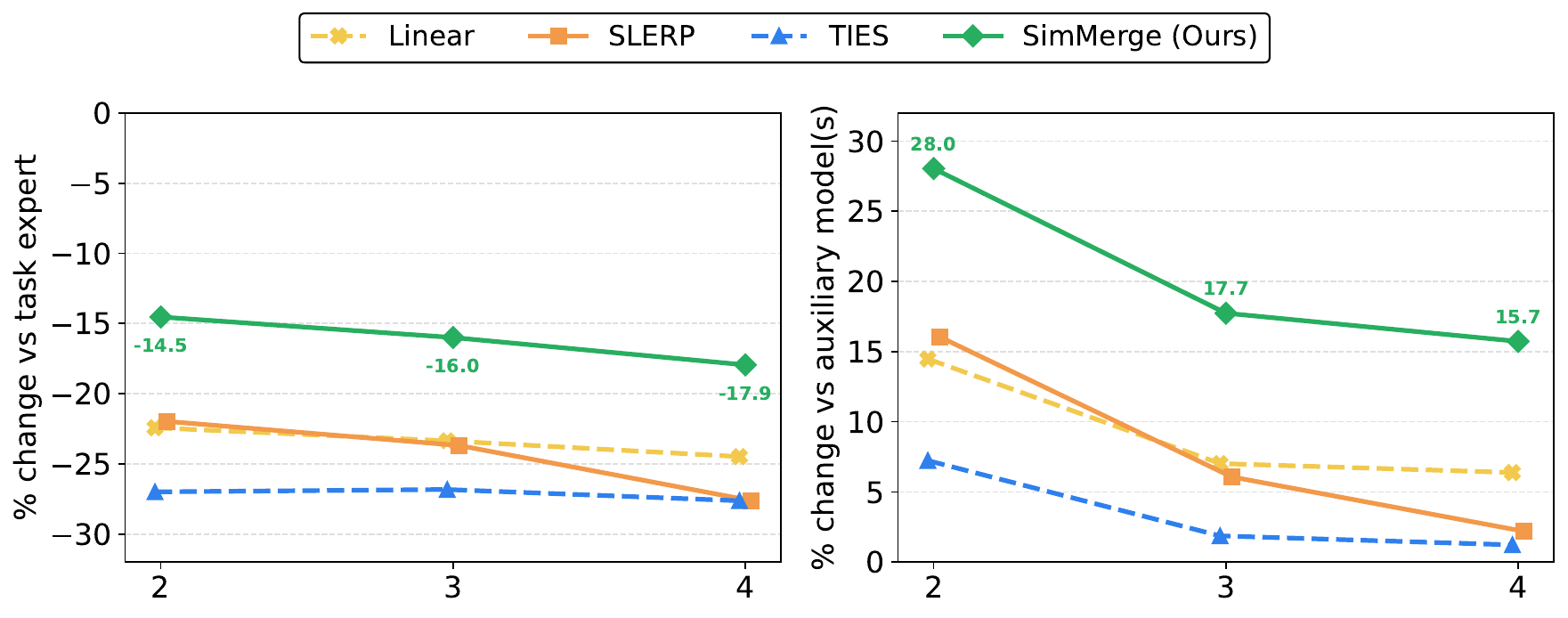}
  \caption{%
    Overall relative performance of 2-, 3-, and 4-way merges, reported as percentage change vs.\ the task expert (left) and vs.\ auxiliary models (right), macro-averaged over all tasks. \simmerge{} consistently improves over auxiliaries while limiting degradation relative to experts as the number of merged models increases.}
  \label{fig:kway-overall}
\end{figure}

We next ask whether a selector trained only on 2-way merges can be lifted to the multiway setting. We encode a $k$-way merge plan $\pi$ as an order-aware sequence of pairwise merges and reuse the same similarity encoder and selector architectures to score candidate plans, without any additional $k$-way supervision.

Figure~\ref{fig:kway-overall} summarizes the effect of increasing the number of merged models from $k=2$ to $3$ and $4$, macro-averaged over all tasks. The \emph{left} panel shows percentage change in performance relative to the task expert, and the \emph{right} panel shows the corresponding change relative to the auxiliary models. 
As expected, increasing $k$ (merging more models by adding more auxiliaries) makes merging more challenging for all methods: degradation relative to the expert grows and gains over auxiliaries shrink. Across all values of $k$, however, \simmerge{} consistently produces the strongest merges among the compared methods, preserving more expert performance while achieving larger improvements over auxiliaries.
For instance, its macro-averaged change relative to the expert is $-14.5\%$, $-16.0\%$, and $-17.9\%$ for $k=2,3,4$, while the change relative to auxiliaries remains positive at $+28.0\%$, $+17.7\%$, and $+15.7\%$, respectively. 
These results mirror the pairwise trends in Section~\ref{sec:results-pairwise} and suggest that similarity-driven selection learned from 2-way merges transfers to multiway merge plans.

Detailed per-task results for 3- and 4-way merges are provided in Appendix~\ref{app:per-task_results}, where we observe the same behavior across code, math, multilingual, and RAG.

\subsection{Effect of Merge Order}
\label{sec:results-order}

For 3-way and 4-way merges, the merge operation is non-associative, so the order in which models are merged can change the final performance. We therefore compare the merge order chosen by our learned selector to a random-order baseline. To isolate the effect of ordering from operator choice, we keep the same sequence of merge operators selected by \simmerge{} fixed and only randomize the merge order. 
\begin{figure}[h!]
  \centering
  \includegraphics[width=0.75\linewidth]{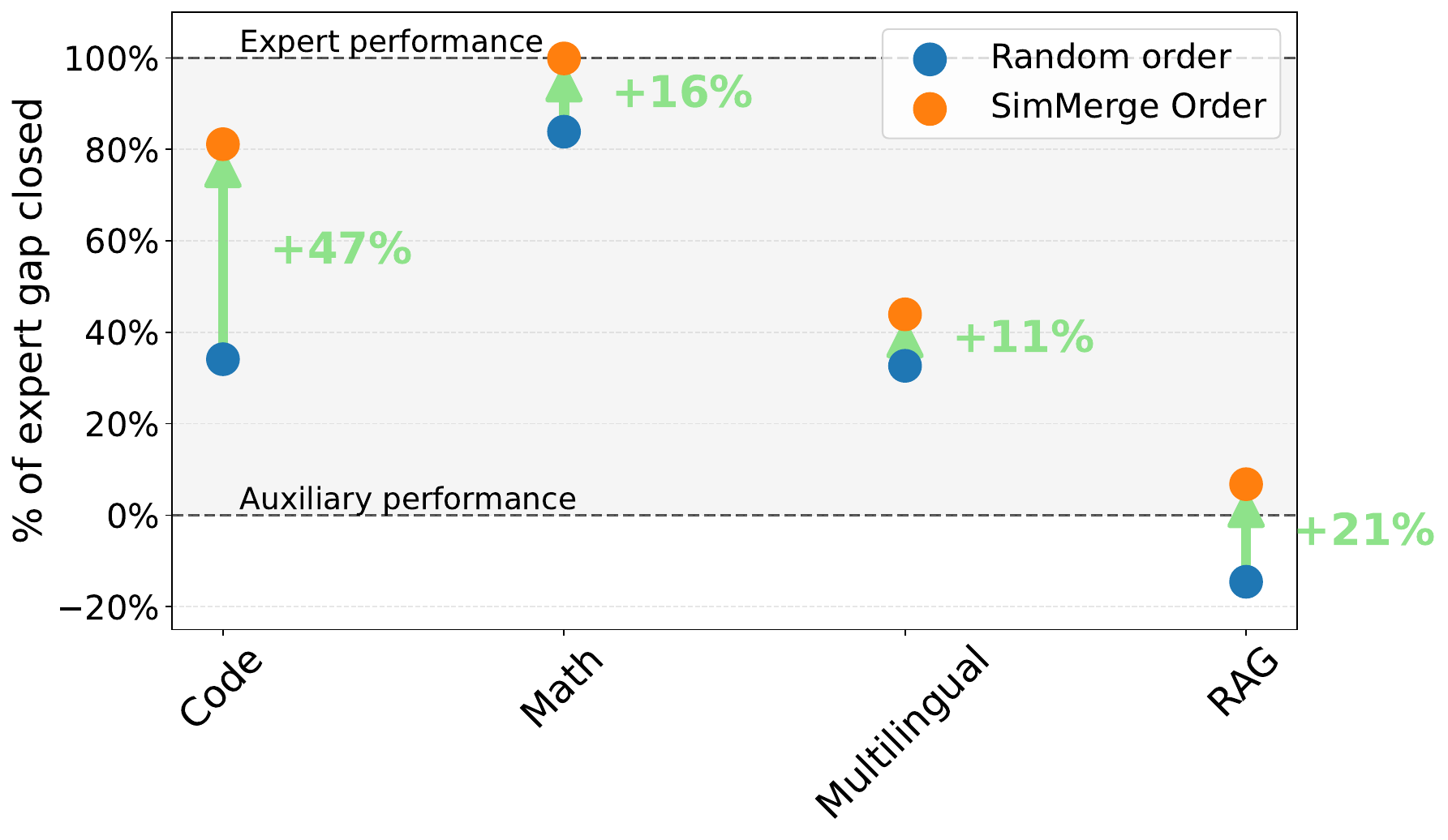}
  \caption{Performance of 3-way merges across Code, Math, Multilingual, and RAG tasks, measured as the percentage of the expert–auxiliary performance gap that is closed. Auxiliary performance corresponds to 0\%, expert performance to 100\%.}
  \label{fig:order-effect-simmerge}
\vspace{-5mm}
\end{figure}

Figure~\ref{fig:order-effect-simmerge} compares \simmerge{}'s selected order to a random-order baseline using the same operator sequence, reported in terms of \emph{GapClosed}. The numeric labels denote the improvement (percentage points) of the learned ordering over the random baseline. Optimized ordering of \simmerge{} delivers consistent gains across tasks, improving \emph{GapClosed} by $+47$ points on Code, $+16$ on Math, $+11$ on Multilingual, and $+21$ on RAG compared to a random ordering. These results indicate that similarity signals are useful not only for selecting merge operators, but also for selecting merge order, with especially large order effects in Code and RAG.

\subsection{Scaling to a 111B-Parameter Model}
\label{sec:results-111b}

To test whether our findings transfer to a substantially larger model, we reuse the selector trained on 7B-parameter checkpoints and repeat the 3-way merging experiments with a 111B-parameter base model, using the same three merge operators and similarity signals as before.

\begin{figure}[h!]
  \centering
  \hspace{-3mm}
  \includegraphics[width=0.90\linewidth]{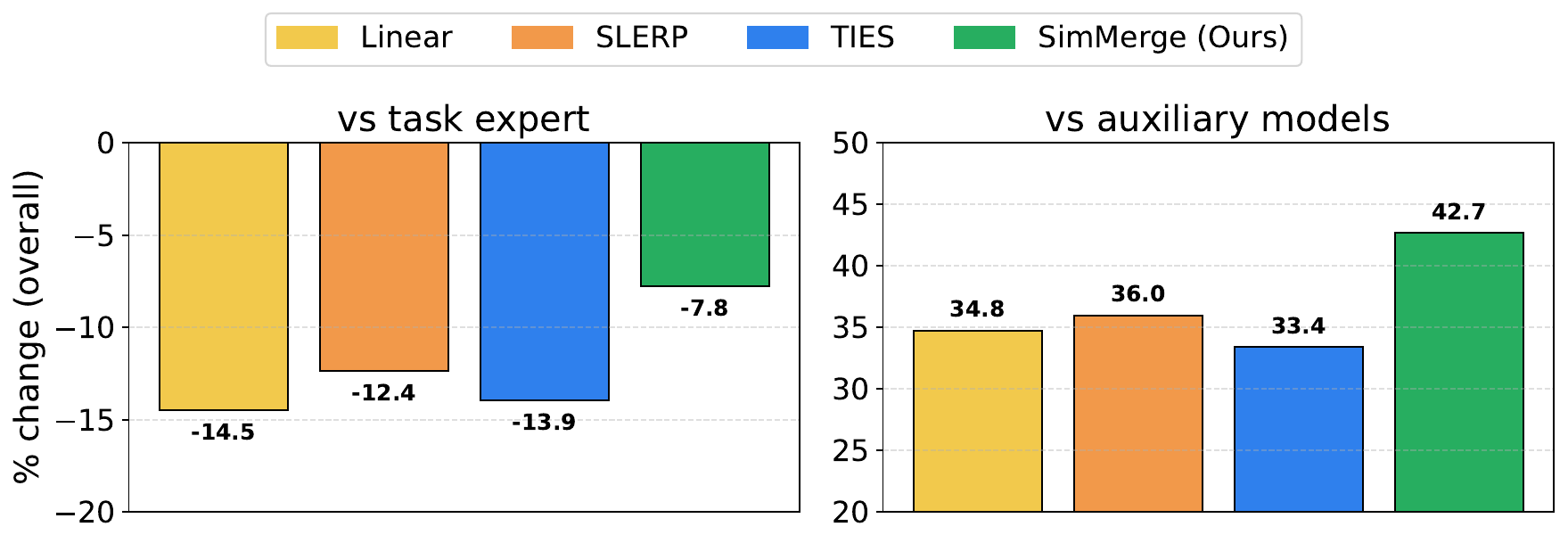}
  \caption{3-way merging at 111B: macro-averaged performance change vs.\ the task expert (left) and auxiliary models (right). \simmerge{} yields the smallest expert degradation and largest gains over auxiliaries, despite training only on 7B checkpoints.}
  \label{fig:111b-overall}
\end{figure}

Figure~\ref{fig:111b-overall} reports the overall percentage change in performance relative to the task expert (left panel) and to the auxiliary baseline (right panel), averaged over tasks.
On the 111B model, \simmerge{} again achieves the best trade-off between expert and auxiliary performance: it reduces degradation relative to the expert to $-7.8$\% compared to $-14.5$\% (\linear), $-12.4$\% (\slerp), and $-13.9$\% (\ties).
At the same time, it achieves the largest improvement over auxiliary models at $+42.7$\% (\simmerge{}) versus $+34.8$\% (\linear), $+36.0$\% (\slerp), and $+33.4$\% (\ties).
Appendix~\ref{app:111b} provides a domain-level breakdown for 3-way merges at 111B using both $\Delta_{\text{expert}}/\Delta_{\text{aux}}$ and \textsc{GapClosed}, and shows the same qualitative pattern across Code, Math, Multilingual, and RAG.
Taken together with the pairwise and multiway results in Sections~\ref{sec:results-pairwise} and~\ref{sec:results-multiway}, these findings demonstrate that predictive merge selection transfers from 7B to 111B parameters without retraining while providing consistent gains over strong fixed merge operators.

\subsection{Online Learning}
\label{sec:bandit-results}

Finally, we evaluate the bandit variant of \simmerge{} in an online setting where the merge operator must be chosen sequentially.
At each round the learner observes similarity features for a 3-way merge configuration including an \emph{instruct} checkpoint in addition to the four main tasks and selects one of \{\linear, \slerp, \ties\}; the reward is the downstream utility of the resulting merge. We compare four policies:
(i) a \emph{uniform random} baseline that chooses operators uniformly at random,
(ii) a \emph{LinUCB} neural-linear upper-confidence-bound policy,
(iii) a \emph{LinTS} neural-linear Thompson-sampling policy (our bandit \simmerge{}), and
(iv) an \emph{oracle} that, for each round, plays the best operator in hindsight given full knowledge of all utilities.
The oracle is not implementable but serves as an upper bound.

\begin{wrapfigure}{r}{0.48\textwidth}
  \centering
  %\vspace{-1mm} % optional: tweak vertical alignment
  \includegraphics[width=0.50\textwidth]{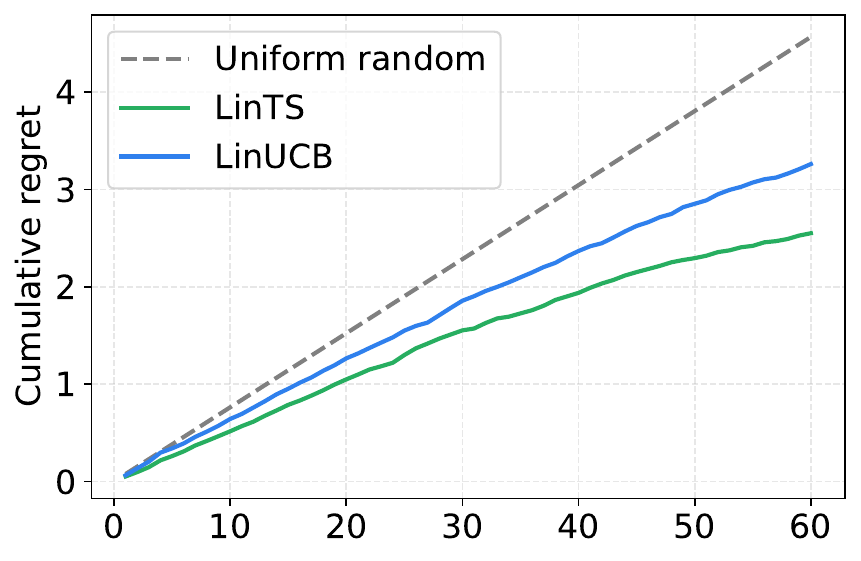}
  \caption{Cumulative regret relative to the oracle over 60 rounds for 3-way merges. Uniform random accumulates regret roughly linearly; both contextual bandits reduce regret quickly, with LinTS consistently lower than LinUCB.}
  \label{fig:bandits_left}
  %\vspace{-1mm} % optional
\end{wrapfigure}

\begin{figure}[h!]
  \centering
  \includegraphics[width=0.90\linewidth]{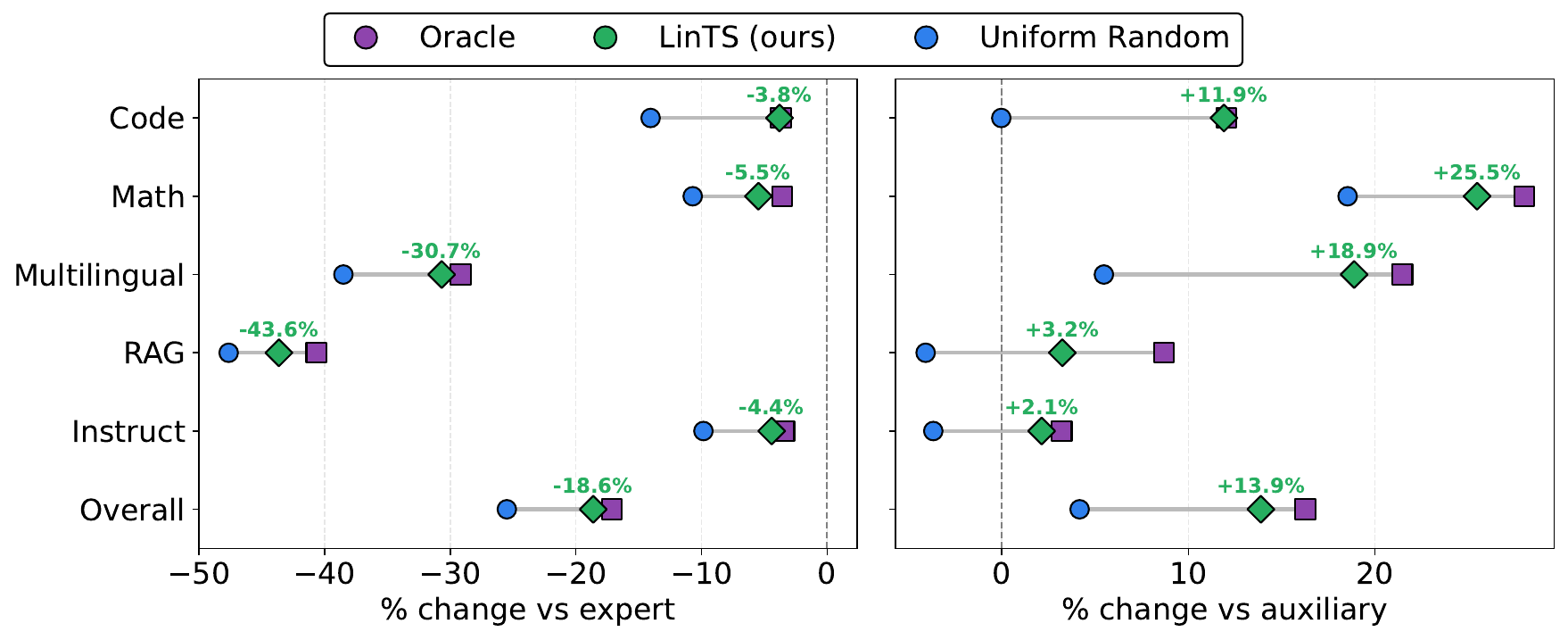}
  \caption{Final percentage change in performance for 3-way merges under uniform random, LinTS, and oracle policies, relative to the task expert (left axis) and auxiliary baseline (right axis), aggregated across domains and macro-average.}
  \label{fig:bandits}
\end{figure}

Figure~\ref{fig:bandits} (left) shows cumulative regret relative to the oracle over 60 rounds. Uniform random accumulates regret roughly linearly. Both contextual bandits learn quickly and substantially reduce regret, but LinTS dominates LinUCB across the entire horizon: it converges faster and tracks the oracle more closely, indicating that the similarity features are informative enough to support efficient exploration and exploitation for operator selection.

To connect regret to downstream quality, Figure~\ref{fig:bandits} (right) reports the final percentage change in performance for three representative policies -- uniform random, LinTS, and oracle -- relative to the task expert and to the auxiliary baseline, aggregated over 3-way merges on Code, Math, Multilingual, RAG, Instruct, and the overall macro-average. LinTS closely tracks the oracle on both axes and consistently improves over uniform random. For example, on the macro-average it reaches $-18.8\%$ relative to the expert while improving by $+13.7\%$ over the auxiliary baseline. These results show that the same similarity features can support a practical online policy via Thompson sampling, enabling adaptation to distribution shift while retaining much of the benefit of full-information operator choice.

\section{Related Work}

\textbf{Weight-space model merging.}
A large body of work studies how to combine models directly in parameter space.
The most common setting assumes experts fine-tuned from a shared base checkpoint and merges them by simple or weighted averaging in weight space or task-vector space~\citep{matena2021fishermerge,ilharcoediting,yadav2023ties}.
Other lines handle models with different initializations via permutation-alignment and subspace-matching techniques~\citep{entezarirole,ainsworth2022gitrebasin}.
Merging has also been explored for parameter-efficient adaptation~\citep{houlsby2019adapters}, LoRA~\citep{hu2022lora}, LEGO~\citep{zhaomerging} and at inference time via soft ensembles or routing~\citep{muqeeth2023soft}.
These methods are largely \emph{operator-centric}: they propose new rules for how to interpolate or combine parameters, often with improved robustness or interference control, but they typically assume that a single merge rule is applied uniformly across all model pairs.

\textbf{Interference-aware and geometry-aware operators.}
Within the shared-checkpoint regime that we also study, several works design more sophisticated merge operators.
Task arithmetic~\citep{ilharcoediting} introduces task vectors and shows that linear combinations can add or negate properties. Fisher-weighted averaging~\citep{matena2022fishermerge} weights parameters by estimated importance from Fisher information. TIES~\citep{yadav2023ties} and DARE~\citep{yu2024language} prune or reweight task vectors to reduce destructive interference before merging, while SLERP~\citep{shoemake1985slerp} follows spherical geodesics in weight or task-vector space.
Recent work also explores structured or alignment-based rules and subspace matching~\citep{tam2023mats,stoica2024zipit,huang2024emrmerging}, and open-source toolkits such as MergeKit~\citep{goddard2024mergekit} have made these operators widely accessible.
In all of these cases, however, the practitioner must still choose which operator, hyperparameters, order or subset of models to apply for a given use case.

\textbf{Automating merge hyperparameters.}
Closer to our goal are methods that tune merge hyperparameters rather than the base operator itself.
Fisher merging~\citep{matena2022fishermerge} uses data-derived Fisher information to compute a weighted average.
AdaMerge~\citep{yangadamerging} chooses task-vector coefficients to minimize merge-model entropy on unlabeled data.
\citet{khalifa2024if} optimize linear merging coefficients over large pools of checkpoints by recycling checkpoints from previous runs to mitigate task tradeoffs, but still rely on repeated merge-and-evaluate iterations.
\citet{su2025fine} propose an automated model merging framework that uses multi-fidelity optimization to search over merge configurations, and introduce fine-grained search spaces such as layer-wise fusion and depth-wise integration to reduce search cost while improving single- and multi-objective performance.
\citet{akiba2025evolutionary} similarly apply evolutionary search to optimize merge recipes, iteratively refining coefficients and layer-wise schedules.
These approaches still operate within a fixed operator family and typically require multiple merge-and-evaluate steps per configuration.

In contrast, our work does not propose a new merge rule or a new way to set its continuous hyperparameters. Instead, we treat \emph{merge selection} itself as the learning problem.
We train a lightweight selector that uses inexpensive model similarity signals to choose between Linear, SLERP, and TIES and to predict the merge order that performs best for a given set of checkpoints on a target task. At test time, \simmerge{} replaces an exhaustive merge-and-evaluate loop with a single forward pass of this selector and a single merge. Unlike prior hyperparameter-search methods, our selector is trained once on 2-way merges and then reused for multiway planning and for larger models, and our neural linear bandit~\citep{riquelme2018deepbayesianbandits} variant supports online adaptation as new tasks, models, or operators are introduced.

\section{Conclusion}

Model merging is a promising alternative to joint training and large ensembles, but its reliability often depends on selecting the merge operator and, for multiway merges, the merge order. We introduced \simmerge{}, a predictive merge selection framework that uses pre-merge similarity signals from model weights and unlabeled probes to choose merge configurations without costly merge-and-evaluate loops.

Across pairwise and multiway merges of 7B checkpoints in code, math, multilingual and retrieval-augmented generation, \simmerge{} consistently outperforms fixed operators by selecting operators per instance and choosing effective merge orders when operators are non-associative. The same selector transfers to multiway merges and 111B models without retraining, improving the expert-auxiliary trade-off while reducing expert degradation. An online neural-linear bandit variant further adapts under partial feedback as tasks and model pools shift.

These results suggest a simple takeaway: reliable model composition is not only about designing better merge rules, but also about learning when to apply the ones we already have. Predictive merge selection turns large checkpoint catalogs into practical, on-demand building blocks under tight evaluation budgets, and it naturally extends as new operators and new model families become available. Future work can expand the operator set, improve intermediate-metric propagation for deeper merge trees, and explore task-agnostic selection that generalizes across evaluation suites with minimal calibration.

\bibliography{main,anthology,addon}

\begin{thebibliography}{37}
\providecommand{\natexlab}[1]{#1}
\providecommand{\url}[1]{\texttt{#1}}
\expandafter\ifx\csname urlstyle\endcsname\relax
  \providecommand{\doi}[1]{doi: #1}\else
  \providecommand{\doi}{doi: \begingroup \urlstyle{rm}\Url}\fi

\bibitem[Abbasi-Yadkori et~al.(2011)Abbasi-Yadkori, P{\'a}l, and Szepesv{\'a}ri]{abbasi2011improved}
Yasin Abbasi-Yadkori, D{\'a}vid P{\'a}l, and Csaba Szepesv{\'a}ri.
\newblock Improved algorithms for linear stochastic bandits.
\newblock In \emph{Advances in Neural Information Processing Systems}, volume~24, 2011.

\bibitem[Agrawal \& Goyal(2013)Agrawal and Goyal]{agrawal2013thompson}
Shipra Agrawal and Navin Goyal.
\newblock Thompson sampling for contextual bandits with linear payoffs.
\newblock In \emph{Proceedings of the 30th International Conference on Machine Learning}, pp.\  127--135. PMLR, 2013.

\bibitem[Ainsworth et~al.(2022)Ainsworth, Hayase, and Srinivasa]{ainsworth2022gitrebasin}
Samuel~K. Ainsworth, Jonathan Hayase, and Siddhartha Srinivasa.
\newblock Git re-basin: Merging models modulo permutation symmetries.
\newblock \emph{arXiv preprint arXiv:2209.04836}, 2022.
\newblock URL \url{https://arxiv.org/abs/2209.04836}.

\bibitem[Akiba et~al.(2025)Akiba, Shing, Tang, Sun, and Ha]{akiba2025evolutionary}
Takuya Akiba, Makoto Shing, Yujin Tang, Qi~Sun, and David Ha.
\newblock Evolutionary optimization of model merging recipes.
\newblock \emph{Nature Machine Intelligence}, 7\penalty0 (2):\penalty0 195--204, 2025.

\bibitem[Boyd \& Vandenberghe(2004)Boyd and Vandenberghe]{boyd2004convex}
Stephen Boyd and Lieven Vandenberghe.
\newblock \emph{Convex Optimization}.
\newblock Cambridge University Press, 2004.

\bibitem[Chen(2021)]{chen2021evaluating}
Mark Chen.
\newblock Evaluating large language models trained on code.
\newblock \emph{arXiv preprint arXiv:2107.03374}, 2021.

\bibitem[Cobbe et~al.(2021)Cobbe, Kosaraju, Bavarian, Chen, Jun, Kaiser, Plappert, Tworek, Hilton, Nakano, Hesse, and Schulman]{cobbe2021training}
Karl Cobbe, Vineet Kosaraju, Mohammad Bavarian, Mark Chen, Heewoo Jun, Lukasz Kaiser, Matthias Plappert, Jerry Tworek, Jacob Hilton, Reiichiro Nakano, Christopher Hesse, and John Schulman.
\newblock Training verifiers to solve math word problems.
\newblock \emph{arXiv preprint arXiv:2110.14168}, 2021.

\bibitem[Cohere et~al.(2025)Cohere, Ahmadian, Ahmed, Alammar, Alizadeh, Alnumay, Althammer, Arkhangorodsky, Aryabumi, Aumiller, et~al.]{cohere2025command}
Team Cohere, Arash Ahmadian, Marwan Ahmed, Jay Alammar, Milad Alizadeh, Yazeed Alnumay, Sophia Althammer, Arkady Arkhangorodsky, Viraat Aryabumi, Dennis Aumiller, et~al.
\newblock Command-{A}: An enterprise-ready large language model.
\newblock \emph{arXiv preprint arXiv:2504.00698}, 2025.

\bibitem[Cover \& Thomas(2006)Cover and Thomas]{cover2006elements}
Thomas~M. Cover and Joy~A. Thomas.
\newblock \emph{Elements of Information Theory}.
\newblock Wiley, 2 edition, 2006.

\bibitem[Csisz{\'a}r(1967)]{csiszar1967information}
Imre Csisz{\'a}r.
\newblock Information-type measures of difference of probability distributions and indirect observation.
\newblock \emph{Studia Sci. Math. Hungarica}, 2:\penalty0 299--318, 1967.

\bibitem[Entezari et~al.(2022)Entezari, Sedghi, Saukh, and Neyshabur]{entezarirole}
Rahim Entezari, Hanie Sedghi, Olga Saukh, and Behnam Neyshabur.
\newblock The role of permutation invariance in linear mode connectivity of neural networks.
\newblock In \emph{International Conference on Learning Representations}, 2022.

\bibitem[Goddard et~al.(2024)Goddard, Siriwardhana, Ehghaghi, Meyers, Karpukhin, Benedict, McQuade, and Solawetz]{goddard2024mergekit}
Charles Goddard, Shamane Siriwardhana, Malikeh Ehghaghi, Luke Meyers, Vlad Karpukhin, Brian Benedict, Mark McQuade, and Jacob Solawetz.
\newblock Arcee's {MergeKit}: A toolkit for merging large language models.
\newblock \emph{arXiv preprint arXiv:2403.13257}, 2024.
\newblock URL \url{https://arxiv.org/abs/2403.13257}.

\bibitem[Hendrycks et~al.(2021)Hendrycks, Burns, Kadavath, Arora, Basart, Tang, Song, and Steinhardt]{hendrycks2021measuring}
Dan Hendrycks, Collin Burns, Saurav Kadavath, Akul Arora, Steven Basart, Eric Tang, Dawn Song, and Jacob Steinhardt.
\newblock Measuring mathematical problem solving with the math dataset.
\newblock \emph{arXiv preprint arXiv:2103.03874}, 2021.

\bibitem[Houlsby et~al.(2019)Houlsby, Giurgiu, Jastrzebski, Morrone, de~Laroussilhe, Gesmundo, Attariyan, and Gelly]{houlsby2019adapters}
Neil Houlsby, Andrei Giurgiu, Stanislaw Jastrzebski, Bruna Morrone, Quentin de~Laroussilhe, Andrea Gesmundo, Mona Attariyan, and Sylvain Gelly.
\newblock Parameter-efficient transfer learning for {NLP}.
\newblock In \emph{International Conference on Machine Learning (ICML)}, volume~97, pp.\  2790--2799. PMLR, 2019.
\newblock URL \url{https://proceedings.mlr.press/v97/houlsby19a.html}.

\bibitem[Hu et~al.(2022)Hu, Shen, Wallis, Allen-Zhu, Li, Wang, Wang, and Chen]{hu2022lora}
Edward~J. Hu, Yelong Shen, Phillip Wallis, Zeyuan Allen-Zhu, Yuanzhi Li, Shean Wang, Lu~Wang, and Weizhu Chen.
\newblock Lora: Low-rank adaptation of large language models.
\newblock In \emph{International Conference on Learning Representations (ICLR)}, 2022.
\newblock URL \url{https://openreview.net/forum?id=nZeVKeeFYf9}.

\bibitem[Huang et~al.(2024)Huang, Yue, et~al.]{huang2024emrmerging}
Chaoyue Huang, Xiangyu Yue, et~al.
\newblock Emr-merging: Tuning-free high-performance model merging.
\newblock In \emph{Advances in Neural Information Processing Systems (NeurIPS)}, 2024.
\newblock URL \url{https://arxiv.org/abs/2405.17461}.

\bibitem[Ilharco et~al.(2023)Ilharco, Ribeiro, Wortsman, Schmidt, Hajishirzi, and Farhadi]{ilharcoediting}
Gabriel Ilharco, Marco~Tulio Ribeiro, Mitchell Wortsman, Ludwig Schmidt, Hannaneh Hajishirzi, and Ali Farhadi.
\newblock Editing models with task arithmetic.
\newblock In \emph{The Eleventh International Conference on Learning Representations}, 2023.

\bibitem[Khalifa et~al.(2024)Khalifa, Tan, Ahmadian, Hosking, Lee, Wang, {\"U}st{\"u}n, Sherborne, and Gall{\'e}]{khalifa2024if}
Muhammad Khalifa, Yi-Chern Tan, Arash Ahmadian, Tom Hosking, Honglak Lee, Lu~Wang, Ahmet {\"U}st{\"u}n, Tom Sherborne, and Matthias Gall{\'e}.
\newblock If you can't use them, recycle them: Optimizing merging at scale mitigates performance tradeoffs.
\newblock \emph{arXiv preprint arXiv:2412.04144}, 2024.

\bibitem[Liu et~al.(2023)Liu, Xia, Wang, and Zhang]{liu2023your}
Jiawei Liu, Chunqiu~Steven Xia, Yuyao Wang, and Lingming Zhang.
\newblock Is your code generated by {ChatGPT} really correct? rigorous evaluation of large language models for code generation.
\newblock \emph{Advances in Neural Information Processing Systems}, 36:\penalty0 21558--21572, 2023.

\bibitem[Matena \& Raffel(2021)Matena and Raffel]{matena2021fishermerge}
Michael Matena and Colin Raffel.
\newblock Merging models with fisher-weighted averaging.
\newblock \emph{arXiv preprint arXiv:2111.09832}, 2021.
\newblock URL \url{https://arxiv.org/abs/2111.09832}.

\bibitem[Matena \& Raffel(2022)Matena and Raffel]{matena2022fishermerge}
Michael Matena and Colin Raffel.
\newblock Merging models with fisher-weighted averaging.
\newblock In \emph{Advances in Neural Information Processing Systems (NeurIPS)}, 2022.
\newblock URL \url{https://arxiv.org/abs/2111.09832}.

\bibitem[Muqeeth et~al.(2023)Muqeeth, Liu, and Raffel]{muqeeth2023soft}
Mohammed Muqeeth, Haokun Liu, and Colin Raffel.
\newblock Soft merging of experts with adaptive routing.
\newblock \emph{arXiv}, abs/2306.03745, 2023.

\bibitem[Patil et~al.(2025)Patil, Mao, Cheng-Jie~Ji, Yan, Suresh, Stoica, and E.~Gonzalez]{patil2025bfcl}
Shishir~G. Patil, Huanzhi Mao, Charlie Cheng-Jie~Ji, Fanjia Yan, Vishnu Suresh, Ion Stoica, and Joseph E.~Gonzalez.
\newblock The berkeley function calling leaderboard (bfcl): From tool use to agentic evaluation of large language models.
\newblock In \emph{Forty-second International Conference on Machine Learning}, 2025.

\bibitem[Riquelme et~al.(2018)Riquelme, Tucker, and Snoek]{riquelme2018deepbayesianbandits}
Carlos Riquelme, George Tucker, and Jasper Snoek.
\newblock Deep bayesian bandits showdown: An empirical comparison of bayesian deep networks for thompson sampling.
\newblock \emph{arXiv preprint arXiv:1802.09127}, 2018.
\newblock URL \url{https://arxiv.org/abs/1802.09127}.

\bibitem[Shi et~al.(2022)Shi, Suzgun, Freitag, Wang, Srivats, Vosoughi, Chung, Tay, Ruder, Zhou, et~al.]{shi2022language}
Freda Shi, Mirac Suzgun, Markus Freitag, Xuezhi Wang, Suraj Srivats, Soroush Vosoughi, Hyung~Won Chung, Yi~Tay, Sebastian Ruder, Denny Zhou, et~al.
\newblock Language models are multilingual chain-of-thought reasoners.
\newblock \emph{arXiv preprint arXiv:2210.03057}, 2022.

\bibitem[Shoemake(1985)]{shoemake1985slerp}
Ken Shoemake.
\newblock Animating rotation with quaternion curves.
\newblock In \emph{SIGGRAPH '85: Proceedings of the 12th Annual Conference on Computer Graphics and Interactive Techniques}, pp.\  245--254. ACM, 1985.
\newblock \doi{10.1145/325165.325242}.
\newblock URL \url{https://dl.acm.org/doi/10.1145/325165.325242}.

\bibitem[Stoica et~al.(2024)Stoica, Bolya, Bjorner, Ramesh, Hearn, and Hoffman]{stoica2024zipit}
George Stoica, Daniel Bolya, Jakob Bjorner, Pratik Ramesh, Taylor Hearn, and Judy Hoffman.
\newblock Zipit! merging models from different tasks without training.
\newblock In \emph{International Conference on Learning Representations (ICLR)}, 2024.
\newblock URL \url{https://openreview.net/forum?id=LEYUkvdUhq}.

\bibitem[Su \& Geiping(2025)Su and Geiping]{su2025fine}
Guinan Su and Jonas Geiping.
\newblock Fine, {I}'ll merge it myself: A multi-fidelity framework for automated model merging.
\newblock \emph{arXiv preprint arXiv:2502.04030}, 2025.

\bibitem[Tam et~al.(2023)Tam, Bansal, and Raffel]{tam2023mats}
Derek Tam, Mohit Bansal, and Colin Raffel.
\newblock Merging by matching models in task parameter subspaces.
\newblock \emph{arXiv preprint arXiv:2312.04339}, 2023.
\newblock URL \url{https://arxiv.org/abs/2312.04339}.

\bibitem[Wortsman et~al.(2022)Wortsman, Ilharco, Gadre, Roelofs, Gontijo-Lopes, Morcos, Namkoong, Farhadi, Carmon, Kornblith, and Schmidt]{wortsman2022modelsoups}
Mitchell Wortsman, Gabriel Ilharco, Samir~Yitzhak Gadre, Rebecca Roelofs, Raphael Gontijo-Lopes, Ari~S. Morcos, Hongseok Namkoong, Ali Farhadi, Yair Carmon, Simon Kornblith, and Ludwig Schmidt.
\newblock Model soups: averaging weights of multiple fine-tuned models improves accuracy without increasing inference time.
\newblock In \emph{International Conference on Machine Learning (ICML)}, volume 162, pp.\  23965--23998. PMLR, 2022.
\newblock URL \url{https://proceedings.mlr.press/v162/wortsman22a.html}.

\bibitem[Yadav et~al.(2023)Yadav, Tam, Choshen, Raffel, and Bansal]{yadav2023ties}
Prateek Yadav, Derek Tam, Leshem Choshen, Colin Raffel, and Mohit Bansal.
\newblock Ties-merging: Resolving interference when merging models.
\newblock \emph{arXiv preprint arXiv:2306.01708}, 2023.
\newblock URL \url{https://arxiv.org/abs/2306.01708}.

\bibitem[Yang et~al.(2024)Yang, Wang, Shen, Liu, Guo, Wang, and Tao]{yangadamerging}
Enneng Yang, Zhenyi Wang, Li~Shen, Shiwei Liu, Guibing Guo, Xingwei Wang, and Dacheng Tao.
\newblock Adamerging: Adaptive model merging for multi-task learning.
\newblock In \emph{The Twelfth International Conference on Learning Representations}, 2024.

\bibitem[Yao et~al.(2025)Yao, Shinn, Razavi, and Narasimhan]{yao2025taubench}
Shunyu Yao, Noah Shinn, Pedram Razavi, and Karthik~R. Narasimhan.
\newblock tau-bench: A benchmark for tool-agent-user interaction in real-world domains.
\newblock In \emph{International Conference on Learning Representations (ICLR)}, 2025.
\newblock URL \url{https://openreview.net/forum?id=roNSXZpUDN}.

\bibitem[Yu et~al.(2024)Yu, Yu, Yu, Huang, and Li]{yu2024language}
Le~Yu, Bowen Yu, Haiyang Yu, Fei Huang, and Yongbin Li.
\newblock Language models are super mario: Absorbing abilities from homologous models as a free lunch.
\newblock In \emph{Forty-first International Conference on Machine Learning}, 2024.

\bibitem[Zeng et~al.(2023)Zeng, Chen, Liu, Jiang, and Jia]{zeng2023mr}
Zhongshen Zeng, Pengguang Chen, Shu Liu, Haiyun Jiang, and Jiaya Jia.
\newblock Mr-gsm8k: A meta-reasoning benchmark for large language model evaluation.
\newblock \emph{arXiv preprint arXiv:2312.17080}, 2023.

\bibitem[Zhao et~al.(2025)Zhao, Shen, Zhu, Li, Su, Wang, and Wu]{zhaomerging}
Ziyu Zhao, Tao Shen, Didi Zhu, Zexi Li, Jing Su, Xuwu Wang, and Fei Wu.
\newblock Merging loras like playing lego: Pushing the modularity of lora to extremes through rank-wise clustering.
\newblock In \emph{The Thirteenth International Conference on Learning Representations}, 2025.

\bibitem[Zhou et~al.(2023)Zhou, Lu, Mishra, Brahma, Basu, Luan, Zhou, and Hou]{zhou2023instruction}
Jeffrey Zhou, Tianjian Lu, Swaroop Mishra, Siddhartha Brahma, Sujoy Basu, Yi~Luan, Denny Zhou, and Le~Hou.
\newblock Instruction-following evaluation for large language models.
\newblock \emph{arXiv preprint arXiv:2311.07911}, 2023.

\end{thebibliography}

\clearpage
\appendix

\section{Merge Operators}
\label{app:merge-operators}

This appendix specifies the merge operators used in the main text: linear interpolation (\linear{}), spherical linear interpolation (\slerp{}), and a TIES-style sign-consistent merge (\ties{}).
All operators act on corresponding parameter tensors of two models $m_a$ and $m_b$ with flattened parameters $\theta_1 = \theta(m_a), \theta_2 = \theta(m_b) \in \mathbb{R}^d$.
Unless stated otherwise, we use a fixed mixing coefficient $\alpha = 0.5$ in the experiments.

\subsection{Linear Interpolation}
\label{app:op-linear}

Linear interpolation (\linear{}) combines parameters by a convex combination,
\[
M_{\text{Lin}}(\theta_1, \theta_2; \alpha)
= (1 - \alpha) \theta_1 + \alpha\theta_2.
\]
In practice we apply this operation layerwise on each parameter tensor such as attention and MLP weights and biases. We do not apply any additional rescaling beyond the convex weights.

\subsection{Spherical Linear Interpolation (SLERP)}
\label{app:op-slerp}

Spherical linear interpolation (\slerp{})~\citep{shoemake1985slerp} interpolates on the unit sphere in parameter space, preserving the norms of the inputs. For each layer we normalize
\[
\hat{\theta}_i
= \frac{\theta_i}{\|\theta_i\|_2}, \quad i \in \{1,2\},
\]
compute the angle $\varphi = \arccos\langle \hat{\theta}_1, \hat{\theta}_2 \rangle$ and form the spherical interpolation
\[
\tilde{\theta}_{\text{unit}}
= \frac{\sin((1-\alpha)\varphi)}{\sin \varphi}\,\hat{\theta}_1
  + \frac{\sin(\alpha \varphi)}{\sin \varphi}\,\hat{\theta}_2.
\]
We then rescale $\tilde{\theta}_{\text{unit}}$ to match the average input norm,
\[
\tilde{\theta}
= \frac{\|\theta_1\|_2 + \|\theta_2\|_2}{2}
  \cdot \frac{\tilde{\theta}_{\text{unit}}}
             {\|\tilde{\theta}_{\text{unit}}\|_2}.
\]
The normalization and rescaling are applied per layer, using the layerwise parameter tensors.
This follows the standard SLERP construction and keeps parameter magnitudes comparable across merges.

\subsection{TIES-Style Sign-Consistent Merge}
\label{app:op-ties}

The TIES-merging (TRIM, ELECT SIGN and MERGE)~\citep{yadav2023ties} is a sign-consistent rule that suppresses conflicting updates while interpolating non-conflicting entries. We abstract it as
\[
M_{\text{TIES}}(\theta_1, \theta_2; \alpha)
= T_\tau(\theta_1, \theta_2; \alpha),
\]
where $\tau \ge 0$ is a threshold hyperparameter.

Let $\theta_1[j], \theta_2[j]$ denote the $j$-th coordinate of the two parameter vectors.
The operator $T_\tau$ is defined coordinate-wise:
\[
T_\tau(\theta_1, \theta_2; \alpha)[j]
=
\begin{cases}
\alpha \,\theta_1[j] + (1-\alpha)\,\theta_2[j],
& \text{if } \theta_1[j]\theta_2[j] > 0
  \text{ and } \max(|\theta_1[j]|, |\theta_2[j]|) \ge \tau, \\[4pt]
\theta_1[j],
& \text{if } \theta_1[j]\theta_2[j] \le 0
  \text{ and } |\theta_1[j]| \ge |\theta_2[j]| \text{ and }
  |\theta_1[j]| \ge \tau, \\[4pt]
\theta_2[j],
& \text{if } \theta_1[j]\theta_2[j] \le 0
  \text{ and } |\theta_2[j]| > |\theta_1[j]| \text{ and }
  |\theta_2[j]| \ge \tau, \\[4pt]
0,
& \text{otherwise.}
\end{cases}
\]
Thus, coordinates with aligned sign and sufficient magnitude are interpolated linearly, while coordinates with sign conflicts are resolved by selecting the larger-magnitude entry, and small-magnitude
coordinates are pruned.
In our implementation, $T_\tau$ is applied layerwise to each parameter tensor, and the same threshold $\tau$ is used across layers. The value of $\tau$ is treated as a hyperparameter and tuned on the pairwise validation split.

\section{Similarity Metrics and Feature Construction}
\label{app:similarity-metrics}

This appendix specifies the similarity metrics and the construction of the feature vectors $x(m_a,m_b,t)$ and $\tilde{x}(m_a,m_b,t)$ used by \simmerge{}.

\subsection{Probe Data and Notation}
\label{app:similarity-probe}

For each task $t \in \mathcal{T}$ we draw an unlabeled probe set
$\mathcal{P}_t = \{x_1, \dots, x_{N_t}\}$ from the input distribution of $t$. No labels are used in any similarity metric.

For a model $m$ and a prompt $x$, let $z_m(x)$ denote the logits produced under teacher forcing. At each decoding position $j$ (token index), we write
\[
p_m(\cdot \mid x,j) = \mathrm{softmax}(z_m(x)_j)
\]
for the next-token predictive distribution over the vocabulary.
For transformer activations, let $h_m^{(\ell)}(x) \in \mathbb{R}^{T \times d_\ell}$ denote the post-residual hidden states at layer $\ell$, where $T$ is the sequence length. When needed, we flatten $h_m^{(\ell)}(x)$ across sequence positions into a single vector in $\mathbb{R}^{T d_\ell}$. Equivalently, we concatenate token representations.

For attention patterns, let $A_m^{(\ell,h)}(x) \in \mathbb{R}^{T \times T}$ denote the attention weight matrix (softmax over keys) at layer $\ell$ and head $h$ for prompt $x$.

For each model $m$ we denote its flattened parameter vector by $\theta(m) \in \mathbb{R}^d$. For brevity, we write $\theta_a=\theta(m_a)$ and $\theta_b=\theta(m_b)$.

\subsection{Data-Based Metrics}
\label{app:similarity-data}

Data-based metrics compare model behavior on the probe set $\mathcal{P}_t$.

\textbf{KL divergence between predictive distributions.}
For an ordered pair $(m_a,m_b)$, prompt $x\in\mathcal{P}_t$, and position $j$, the pointwise KL divergence is
\[
D_{\mathrm{KL}}\!\big(p_a(\cdot \mid x,j)\,\|\,p_b(\cdot \mid x,j)\big)
= \sum_{i} p_a(i \mid x,j)\,
  \log \frac{p_a(i \mid x,j)}{p_b(i \mid x,j)}.
\]
We average over positions and prompts to obtain
\[
\mathrm{KL}_{\text{mean}}(m_a,m_b,t)
=
\frac{1}{N_t}\sum_{x\in\mathcal{P}_t}\;\frac{1}{|J(x)|}\sum_{j\in J(x)}
D_{\mathrm{KL}}\!\big(p_a(\cdot \mid x,j)\,\|\,p_b(\cdot \mid x,j)\big),
\]
where $J(x)$ is the set of teacher-forced positions used for evaluation.
We additionally record robust summary statistics over prompts (median and empirical quantiles, such as 25th/75th/90th percentiles).
KL is computed in log-space with standard numerical stabilization.

\textbf{Activation cosine similarity.}
For each layer $\ell$ and prompt $x\in\mathcal{P}_t$, let $\mathrm{vec}(h_m^{(\ell)}(x))\in\mathbb{R}^{Td_\ell}$ denote the flattened hidden states. We define
\[
\cos_h^{(\ell)}(m_a,m_b,x)
=
\frac{\left\langle \mathrm{vec}(h_a^{(\ell)}(x)),\, \mathrm{vec}(h_b^{(\ell)}(x)) \right\rangle}
{\left\|\mathrm{vec}(h_a^{(\ell)}(x))\right\|_2 \left\|\mathrm{vec}(h_b^{(\ell)}(x))\right\|_2},
\qquad
\cos_h^{(\ell)}(m_a,m_b,t)
=
\frac{1}{N_t}\sum_{x\in\mathcal{P}_t}\cos_h^{(\ell)}(m_a,m_b,x).
\]
We keep either the full per-layer sequence $\{\cos_h^{(\ell)}\}_{\ell=1}^L$ for later summarization or aggregate across layers immediately.

\textbf{Attention-pattern cosine similarity.}
For each layer $\ell$, head $h$, and prompt $x\in\mathcal{P}_t$, we flatten attention matrices and compute
\[
\cos_A^{(\ell,h)}(m_a,m_b,x)
=
\frac{\left\langle \mathrm{vec}(A_a^{(\ell,h)}(x)),\, \mathrm{vec}(A_b^{(\ell,h)}(x)) \right\rangle}
{\left\|\mathrm{vec}(A_a^{(\ell,h)}(x))\right\|_2 \left\|\mathrm{vec}(A_b^{(\ell,h)}(x))\right\|_2}.
\]
We summarize these values across prompts, heads, and layers using the same robust statistics as above (mean/median/quantiles), yielding a compact set of attention-similarity features.

\subsection{Weight-Based Metrics}
\label{app:similarity-weights}

Weight-based metrics compare parameters directly and do not depend on $\mathcal{P}_t$.

\textbf{Weight cosine similarity.}
For flattened parameter vectors $\theta_a,\theta_b\in\mathbb{R}^d$,
\[
\cos_W(m_a,m_b)
=
\frac{\langle \theta_a, \theta_b \rangle}{\|\theta_a\|_2\,\|\theta_b\|_2}.
\]
We optionally compute layerwise or module-restricted variants by restricting $\theta_a,\theta_b$ to parameters of a given transformer block or to attention/MLP submodules.

\textbf{Weight $\ell_2$ distance.}
\[
d_W(m_a,m_b)=\|\theta_a-\theta_b\|_2,
\]
again optionally computed per-layer or per-module by restriction to parameter subsets.

\textbf{Weight norms.}
We record $\|\theta_a\|_2$ and $\|\theta_b\|_2$ (and optionally their layerwise/modulewise norms) to capture global scale differences that can interact with merge behavior.

\subsection{Feature Vector Construction}
\label{app:similarity-features-construct}

Each metric yields either a scalar or a short sequence indexed by layers and, for attention, optionally heads. To obtain a fixed-dimensional representation, sequence-valued metrics are summarized using robust statistics such as the mean, median, and selected quantiles, and all summaries are concatenated into a single feature vector
\[
x(m_a,m_b,t)\in\mathbb{R}^m.
\]

By default, we append an explicit task encoding $c(t)\in\mathbb{R}^{d_c}$ and use
\[
\tilde{x}(m_a,m_b,t) = x(m_a,m_b,t)\oplus c(t)\in\mathbb{R}^{m+d_c}
\]
as the input to all learned components. We also evaluate a task-agnostic variant that omits $c(t)$; the comparison is reported in Appendix~\ref{app:classifier-acc}. The improvement from the task encoding is modest but consistent, so we keep it enabled in the main experiments.

\section{Why Pairwise Training Can Transfer to Multi-Way Planning}
\label{app:multiway-theory}

This section provides intuition for why a scorer trained using pairwise-derived signals can be effective for ranking multi-way merge plans.
For a $k$-way plan $\pi=(m_{i_1}\!\rightarrow\!\cdots\!\rightarrow\!m_{i_k})$ on task $t$, \simmerge{} represents the plan by concatenating step-wise feature blocks,
\[
X(\pi,t)=\big[x(m_{i_1},m_{i_2},t),\,\ldots,\,x(m_{i_{k-1}},m_{i_k},t)\big]\oplus c(t).
\]
This construction is motivated by the observation that non-associativity makes the \emph{local} interaction between consecutive merge steps consequential: changing the order changes which pairs interact early versus late, and these interactions are reflected in the corresponding pairwise similarity regimes.

A sufficient condition for this representation to be useful is that the utility of executing a plan, $U(\pi,t)$, depends smoothly on (or can be well-approximated by) a low-order function of these step-wise interactions. For example, if
\[
U(\pi,t)\approx \sum_{s=1}^{k-1}\psi\big(x(m_{i_s},m_{i_{s+1}},t),\,t\big)
\]
for some unknown function $\psi$, then a learned plan scorer can estimate $U(\pi,t)$ from $X(\pi,t)$ by aggregating step-wise contributions.
More generally, if $U(\pi,t)=F(x_1,\ldots,x_{k-1},c(t))$ is a sufficiently smooth function of the step blocks $x_s$, then a first-order expansion around typical interaction regimes yields an approximately additive dependence on the concatenated features.
This motivates learning $f_{\mathrm{plan}}$ on plan representations built from the same pairwise feature blocks used for operator selection.

\section{Propagation of Similarity Metrics to Multi-Way Plans}
\label{app:metric-propagation}

This appendix details how we construct approximate similarity features for intermediate steps when scoring multi-way merge plans, without explicitly constructing and evaluating intermediate merged parameters for each candidate plan.

A multi-way plan involves intermediate merged checkpoints. To score candidate plans efficiently, we use a \emph{proxy} representation for an intermediate step formed by merging $a$ and $b$ with coefficient $\alpha$.
In the main experiments we fix $\alpha=\tfrac12$ and treat the intermediate as an equal-weight combination for the purpose of constructing features.
For data-based quantities such as KL, where the intermediate model's predictive distribution is not available without executing the merge, we use mixture-inspired proxy estimates derived from standard inequalities; these values serve as inexpensive features rather than exact measurements of the true intermediate model.

Orientation is explicit because some metrics are asymmetric (notably KL). We write
\[
(a{+}b,\,c): \quad G_L=(1-\alpha)P_a+\alpha P_b,\; G_R=P_c,
\quad\text{and}\quad
(c,\,a{+}b): \quad G_L=P_c,\; G_R=(1-\alpha)Q_a+\alpha Q_b,
\]
where $P_\cdot$ and $Q_\cdot$ denote predictive distributions on the probe set or distributional proxies derived from logits.

\subsection{KL Divergence Proxies}

By the log-sum inequality and the joint convexity of KL (more generally, $f$-divergences) in each argument \citep{csiszar1967information,cover2006elements},
\begin{align}
\mathrm{KL}\big((1-\alpha)P_a + \alpha P_b \,\|\, P_c\big)
&\le (1-\alpha)\,\mathrm{KL}(P_a \,\|\, P_c) + \alpha\,\mathrm{KL}(P_b \,\|\, P_c), \label{eq:kl-left}\\
\mathrm{KL}\big(P_c \,\|\, (1-\alpha)Q_a + \alpha Q_b\big)
&\le (1-\alpha)\,\mathrm{KL}(P_c \,\|\, Q_a) + \alpha\,\mathrm{KL}(P_c \,\|\, Q_b). \label{eq:kl-right}
\end{align}

More generally, when both arguments are mixtures,
\[
G_L := \sum_i w_i P_i,
\qquad
G_R := \sum_j v_j Q_j,
\]
with $w_i, v_j \ge 0$ and $\sum_i w_i=\sum_j v_j=1$, we have
\[
\mathrm{KL}(G_L \,\|\, G_R) \le \sum_{i,j} w_i v_j\, \mathrm{KL}(P_i \,\|\, Q_j).
\]

In practice, we use the right-hand sides of \eqref{eq:kl-left} and \eqref{eq:kl-right} as propagated \emph{proxy} values.

\subsection{$\ell_2$ Parameter Distance Proxies}

Let $\theta_a,\theta_b,\theta_c\in\mathbb{R}^d$ be flattened parameter vectors and $L_2(\theta,\theta')=\|\theta-\theta'\|_2$.
By the triangle inequality and positive homogeneity of norms \citep{boyd2004convex},
\begin{align}
\big\| (1-\alpha)\theta_a + \alpha \theta_b - \theta_c \big\|_2
&\le (1-\alpha)\,\|\theta_a-\theta_c\|_2 + \alpha\,\|\theta_b-\theta_c\|_2, \label{eq:l2-left}\\
\big\| \theta_c - \big((1-\alpha)\theta_a + \alpha \theta_b\big) \big\|_2
&\le (1-\alpha)\,\|\theta_c-\theta_a\|_2 + \alpha\,\|\theta_c-\theta_b\|_2. \label{eq:l2-right}
\end{align}
We use the right-hand sides as propagated proxy values and mark them \textit{proxy upper}.

\subsection{Cosine Similarity Proxies (Weights or Attention Patterns)}

Define $\cos(u,v)=\frac{\langle u,v\rangle}{\|u\|_2\,\|v\|_2}$.
Because the denominator is nonlinear in mixtures, an exact cosine with an intermediate proxy would require dot products and norms that are typically not logged for every possible intermediate.
We therefore use a simple, stable proxy:
\begin{equation}
\cos\big((1-\alpha)u_a + \alpha u_b,\; u_c\big)
\approx (1-\alpha)\,\cos(u_a,u_c) + \alpha\,\cos(u_b,u_c),
\label{eq:cos-heur}
\end{equation}
and similarly for $\cos\big(u_c,\,(1-\alpha)u_a+\alpha u_b\big)$.
We clip the resulting values to $[-1,1]$.
This rule is used both for weight-vector cosines (with $u_\cdot=\theta_\cdot$) and for attention-pattern cosines (with $u_\cdot=\mathrm{vec}(A_\cdot^{(\ell,h)}(x))$ after summarization).

These propagated proxy values are aggregated using the same robust statistics as in Appendix~\ref{app:similarity-metrics} and inserted into the plan representation $X(\pi,t)$ whenever a similarity involving an intermediate step is required.
This allows \simmerge{} to score candidate multi-step merge sequences using a precomputed pairwise similarity table, without recomputing similarities for every hypothetical intermediate merge.

\section{Evaluation Benchmarks and Metrics}
\label{app:evals}

We provide additional details on the benchmarks and evaluation metrics used for each task domain.

\textbf{Math reasoning.} We evaluate mathematical reasoning on \emph{MATH}~\citep{hendrycks2021measuring} and \emph{GSM8K}~\citep{cobbe2021training}. Both benchmarks consist of grade-school to competition-level math problems that require multi-step reasoning and symbolic manipulation. Models are evaluated using \emph{exact match} accuracy, where a prediction is considered correct only if the final answer exactly matches the reference solution. For GSM8K, answers are normalized following standard evaluation protocols to account for formatting differences.

\textbf{Multilingual question answering.}
Multilingual performance is measured using an internal multilingual QA suite together with \emph{MGSM}~\citep{shi2022language}, which extends GSM-style math reasoning to multiple languages.
MGSM evaluates cross-lingual generalization and reasoning robustness.
Performance is reported using accuracy and win-rate metrics, where win-rate measures the fraction of examples on which a model’s answer is preferred over a baseline under automatic or human evaluation, depending on the benchmark.
These metrics capture both correctness and relative answer quality across languages.

\textbf{Code generation.} Code generation is evaluated on \emph{HumanEval\_Python}~\citep{chen2021evaluating} and \emph{MBPP+}~\citep{liu2023your}.
Both benchmarks assess functional correctness of generated programs against unit tests. We report \emph{pass@1}, which measures the probability that the first generated solution passes all test cases.
This metric reflects single-sample code generation quality and is standard in code evaluation.

\textbf{Retrieval-augmented generation (RAG).}
RAG performance is evaluated on \emph{TauBench}~\citep{yao2025taubench} and
\emph{BFCL}~\citep{patil2025bfcl}, which test a model's ability to integrate retrieved evidence into accurate responses. We report accuracy and F1 score, depending on the benchmark, following their official evaluation protocols.
These metrics assess both answer correctness and overlap with reference responses, capturing retrieval grounding quality.

\textbf{Instruction following.} For instruction-following experiments used in the bandit setting, we evaluate on \emph{IFEval}~\citep{zhou2023instruction}.
IFEval measures a model's ability to follow explicit instructions and constraints.
Performance is reported using the benchmark's standard instruction-compliance score, which aggregates binary success indicators across multiple instruction types.

All evaluations are run three times with different random seeds, and we report the mean score. This reduces variance due to stochastic decoding and ensures stable comparisons across merge methods.

\section{Additional Results}
\label{app:addtional_results}

\subsection{Classifier Accuracy and Task-Encoding Ablation}
\label{app:classifier-acc}

To quantify predictive accuracy, we report confusion matrices for the offline selector on the held-out pairwise test set of 60 merges.
We compare our default task-conditioned representation, which appends a task encoding $c(t)$ to the similarity features, against a task-agnostic variant that omits $c(t)$.

\begin{figure}[h!]
  \centering
  \begin{subfigure}{0.49\linewidth}
    \centering
    \includegraphics[width=\linewidth]{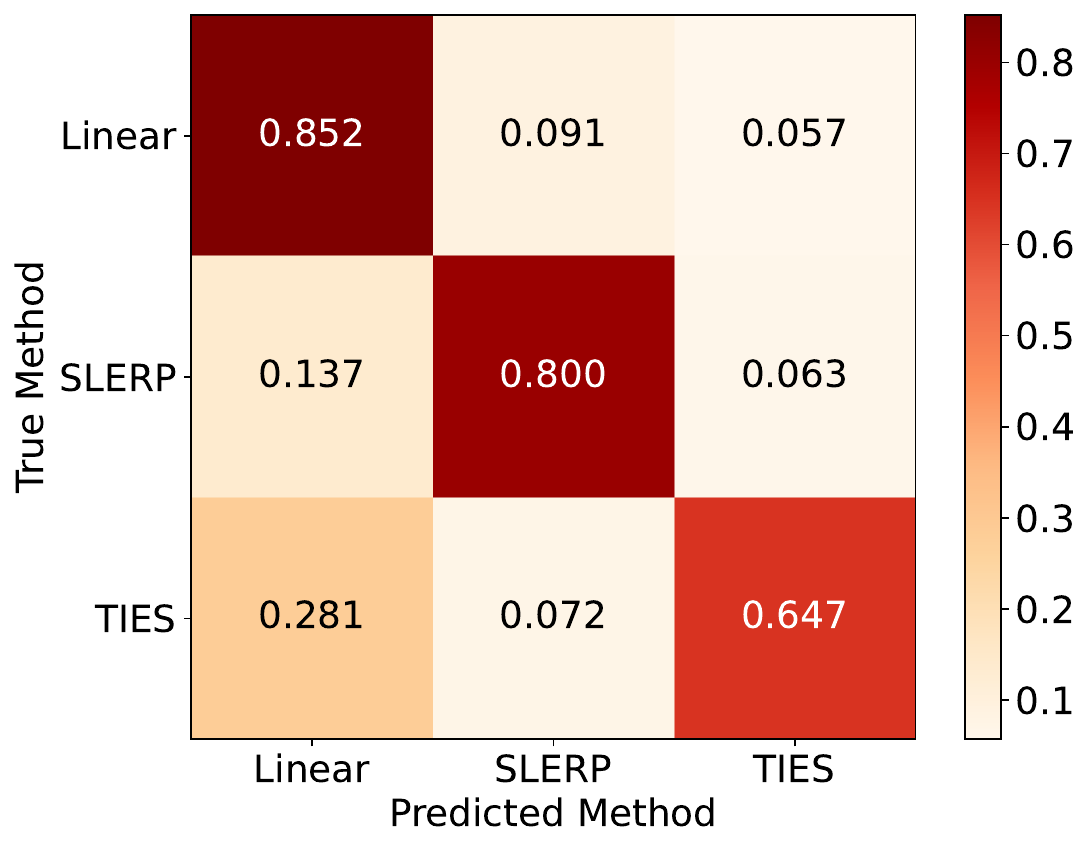}
    \caption{Without task encoding}
  \end{subfigure}
  \begin{subfigure}{0.49\linewidth}
    \centering
    \includegraphics[width=\linewidth]{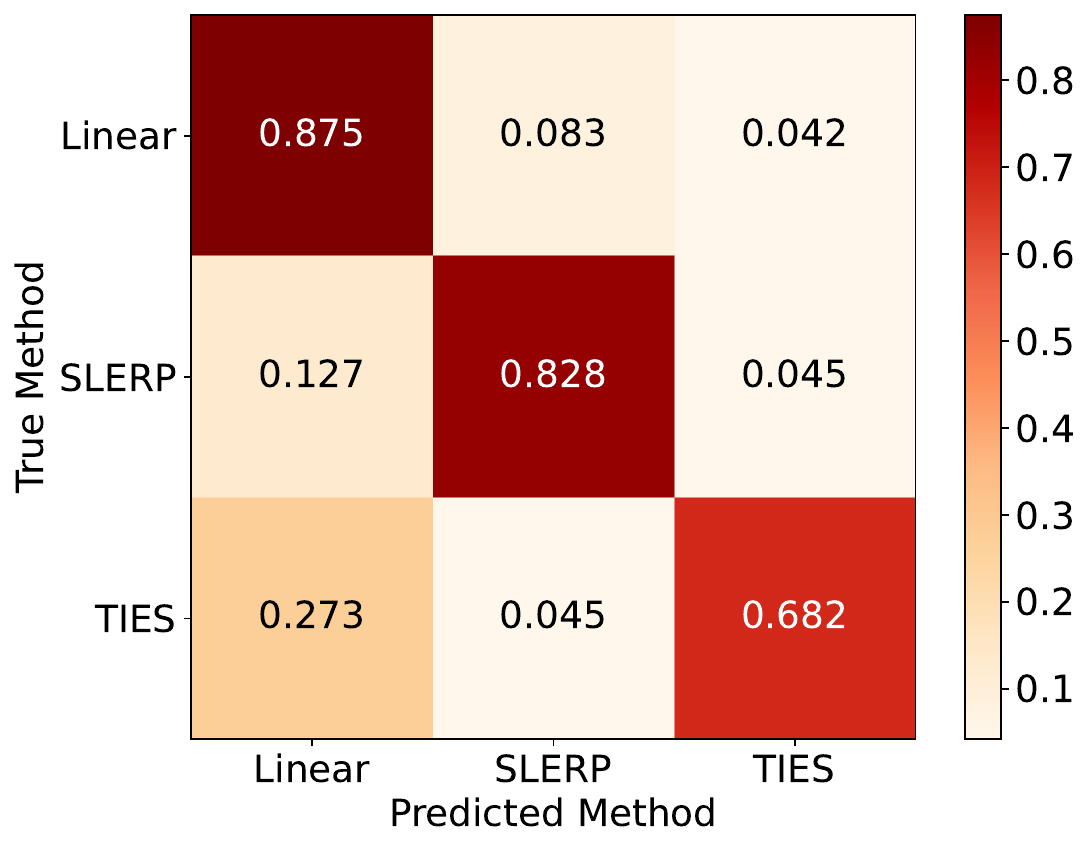}
    \caption{With task encoding}
  \end{subfigure}
  \caption{Confusion matrices of the offline selector on the held-out pairwise test set.
  Appending a task encoding improves per-class accuracy from 85.2\% to 87.5\% for Linear, from 80.0\% to 82.8\% for SLERP, and from 64.7\% to 68.2\% for TIES.}
  \label{fig:pairwise-confusion-ablation}
\end{figure}

Figure~\ref{fig:pairwise-confusion-ablation} shows that task conditioning yields a small but consistent improvement across all classes.
With the task encoding, the selector correctly identifies Linear in 87.5\% of cases, SLERP in 82.8\%, and TIES in 68.2\%.
Without the task encoding, accuracy drops to 85.2\% for Linear, 80.0\% for SLERP, and 64.7\% for TIES.
Across both settings, most errors occur when the true operator is TIES, reflecting that TIES occupies a narrower regime and is easier to confuse with Linear or SLERP.
Overall, these results support that similarity features capture the relationships that drive operator preference, and that a lightweight task encoding provides an additional, modest gain.

\subsection{Detailed Per-Task Results Across Merge Sizes}
\label{app:per-task_results}

Tables~\ref{tab:pairwise-summary}–\ref{tab:quad-summary} provide detailed per-task summaries for pairwise, three-way, and four-way merges of 7B checkpoints. For each task and merge method, we report the mean task performance, absolute differences from the expert and auxiliary baselines, and the corresponding relative changes. Highlighting the best fixed operator per task makes explicit how the strongest baseline varies across settings, while bolding \simmerge{} emphasizes its consistent advantage across tasks and merge sizes.

\begin{table}[h!]
\caption{Pairwise (2-way) merges: per-task performance summary. Mean is the task-level average score. Diff.Exp and Diff.Aux denote absolute differences from the expert and auxiliary baselines, \%Exp and \%Aux are the corresponding relative changes. The best fixed operator (among Linear, SLERP, and TIES) is highlighted in blue for each task, \simmerge{} is bold.}
\centering
\small
\setlength{\tabcolsep}{4.5pt}
\begin{tabular}{l l r r r r r}
\toprule
Task & Method & Mean & Diff.Exp & Diff.Aux & \%Exp & \%Aux \\
\midrule
Code & \fixedbest{Linear} & \fixedbest{0.60} & \fixedbest{-0.03} & \fixedbest{0.07} & \fixedbest{-4.70} & \fixedbest{12.31} \\
     & SLERP & 0.59 & -0.04 & 0.05 & -7.07 & 9.52 \\
     & TIES & 0.56 & -0.07 & 0.02 & -11.66 & 4.11 \\
     & \textbf{\simmerge{}} & \textbf{0.62} & \textbf{-0.02} & \textbf{0.08} & \textbf{-2.40} & \textbf{15.02} \\
\midrule
Math & Linear & 0.68 & -0.07 & 0.11 & -9.15 & 19.20 \\
     & SLERP & 0.70 & -0.05 & 0.13 & -6.61 & 22.53 \\
     & \fixedbest{TIES} & \fixedbest{0.72} & \fixedbest{-0.03} & \fixedbest{0.15} & \fixedbest{-3.98} & \fixedbest{25.98} \\
     & \textbf{\simmerge{}} & \textbf{0.74} & \textbf{-0.01} & \textbf{0.16} & \textbf{-1.90} & \textbf{28.80} \\
\midrule
Multilingual & \fixedbest{Linear} & \fixedbest{0.44} & \fixedbest{-0.16} & \fixedbest{0.09} & \fixedbest{-26.40} & \fixedbest{26.77} \\
            & SLERP & 0.38 & -0.22 & 0.03 & -36.55 & 9.29 \\
            & TIES & 0.32 & -0.28 & -0.03 & -46.24 & -7.40 \\
            & \textbf{\simmerge{}} & \textbf{0.46} & \textbf{-0.13} & \textbf{0.12} & \textbf{-22.27} & \textbf{33.89} \\
\midrule
RAG & Linear & 0.21 & -0.20 & -0.00 & -49.46 & -0.38 \\
    & \fixedbest{SLERP} & \fixedbest{0.25} & \fixedbest{-0.15} & \fixedbest{0.05} & \fixedbest{-37.65} & \fixedbest{22.88} \\
    & TIES & 0.22 & -0.19 & 0.01 & -46.09 & 6.26 \\
    & \textbf{\simmerge{}} & \textbf{0.28} & \textbf{-0.13} & \textbf{0.07} & \textbf{-32.11} & \textbf{33.81} \\
\bottomrule
\end{tabular}
\label{tab:pairwise-summary}
\end{table}

In the pairwise setting (Table~\ref{tab:pairwise-summary}), the best fixed operator differs substantially across tasks: Linear performs best on Code and Multilingual, TIES on Math, and SLERP on RAG. This variability confirms that no single merge operator dominates even in the simplest two-model regime. Across all tasks, \simmerge{} consistently achieves higher mean performance than the best fixed operator, simultaneously reducing degradation relative to the expert and improving more over the auxiliary baseline. These results establish that similarity features are predictive of operator choice and motivate learning instance-specific merge decisions.

\begin{table}[h!]
\caption{Three-way (k=3) merges: per-task performance summary using the same metrics as Table~\ref{tab:pairwise-summary}. The best fixed operator (among Linear, SLERP, and TIES) is highlighted in blue for each task; \simmerge{} is bold.}
\centering
\small
\setlength{\tabcolsep}{4.5pt}
\begin{tabular}{l l r r r r r}
\toprule
Task & Method & Mean & Diff.Exp & Diff.Aux & \%Exp & \%Aux \\
\midrule
Code & \fixedbest{Linear} & \fixedbest{0.60} & \fixedbest{-0.03} & \fixedbest{0.06} & \fixedbest{-5.39} & \fixedbest{11.77} \\
     & SLERP & 0.57 & -0.06 & 0.04 & -9.18 & 7.29 \\
     & TIES  & 0.51 & -0.12 & -0.02 & -18.68 & -3.94 \\
     & \textbf{\simmerge{}} & \textbf{0.61} & \textbf{-0.02} & \textbf{0.08} & \textbf{-2.90} & \textbf{14.71} \\
\midrule
Math & Linear & 0.64 & -0.09 & 0.04 & -12.48 & 7.33 \\
     & SLERP & 0.70 & -0.03 & 0.11 & -3.89 & 17.86 \\
     & \fixedbest{TIES} & \fixedbest{0.71} & \fixedbest{-0.02} & \fixedbest{0.11} & \fixedbest{-2.76} & \fixedbest{19.25} \\
     & \textbf{\simmerge{}} & \textbf{0.73} & \textbf{-0.00} & \textbf{0.13} & \textbf{-0.02} & \textbf{22.61} \\
\midrule
Multilingual & \fixedbest{Linear} & \fixedbest{0.42} & \fixedbest{-0.15} & \fixedbest{0.08} & \fixedbest{-25.79} & \fixedbest{22.82} \\
            & SLERP & 0.36 & -0.21 & 0.02 & -36.29 & 5.45 \\
            & TIES  & 0.34 & -0.23 & -0.00 & -39.76 & -0.30 \\
            & \textbf{\simmerge{}} & \textbf{0.44} & \textbf{-0.13} & \textbf{0.10} & \textbf{-22.20} & \textbf{28.77} \\
\midrule
RAG & Linear & 0.20 & -0.20 & -0.03 & -49.76 & -13.81 \\
    & \fixedbest{SLERP} & \fixedbest{0.22} & \fixedbest{-0.18} & \fixedbest{-0.01} & \fixedbest{-45.38} & \fixedbest{-6.30} \\
    & TIES  & 0.21 & -0.18 & -0.02 & -46.09 & -7.52 \\
    & \textbf{\simmerge{}} & \textbf{0.24} & \textbf{-0.15} & \textbf{0.01} & \textbf{-38.88} & \textbf{4.85} \\
\bottomrule
\end{tabular}
\label{tab:triple-summary}
\end{table}

For three-way merges (Table~\ref{tab:triple-summary}), overall performance decreases relative to the pairwise setting, reflecting the increased difficulty of composing multiple models. Nevertheless, the same qualitative patterns persist: the identity of the strongest fixed operator remains task-dependent, and fixed baselines occasionally fail to improve over auxiliaries, particularly on RAG and Multilingual. In contrast, \simmerge{} consistently yields the highest mean performance across all tasks, incurring the smallest expert degradation while maintaining positive gains over auxiliary models in every domain.

\begin{table}[h!]
\caption{Four-way (k=4) merges: per-task performance summary using the same metrics as Table~\ref{tab:pairwise-summary}. The best fixed operator (among Linear, SLERP, and TIES) is highlighted in blue for each task; \simmerge{} is bold.}
\centering
\small
\setlength{\tabcolsep}{4.5pt}
\begin{tabular}{l l r r r r r}
\toprule
Task & Method & Mean & Diff.Exp & Diff.Aux & \%Exp & \%Aux \\
\midrule
Code & Linear & 0.53 & -0.08 & 0.01 & -13.30 & 1.10 \\
     & SLERP  & 0.51 & -0.10 & -0.02 & -16.89 & -3.09 \\
     & \fixedbest{TIES} & \fixedbest{0.53} & \fixedbest{-0.08} & \fixedbest{0.01} & \fixedbest{-13.00} & \fixedbest{1.44} \\
     & \textbf{\simmerge{}} & \textbf{0.59} & \textbf{-0.02} & \textbf{0.07} & \textbf{-3.36} & \textbf{12.68} \\
\midrule
Math & Linear & 0.69 & -0.03 & 0.15 & -4.70 & 27.01 \\
     & SLERP  & 0.68 & -0.04 & 0.14 & -5.54 & 25.67 \\
     & \fixedbest{TIES} & \fixedbest{0.71} & \fixedbest{-0.02} & \fixedbest{0.17} & \fixedbest{-2.09} & \fixedbest{31.59} \\
     & \textbf{\simmerge{}} & \textbf{0.71} & \textbf{-0.02} & \textbf{0.17} & \textbf{-2.09} & \textbf{31.59} \\
\midrule
Multilingual & \fixedbest{Linear} & \fixedbest{0.38} & \fixedbest{-0.16} & \fixedbest{0.04} & \fixedbest{-29.95} & \fixedbest{11.54} \\
            & SLERP  & 0.33 & -0.22 & -0.01 & -39.72 & -4.01 \\
            & TIES   & 0.29 & -0.26 & -0.05 & -47.19 & -15.91 \\
            & \textbf{\simmerge{}} & \textbf{0.40} & \textbf{-0.14} & \textbf{0.06} & \textbf{-25.63} & \textbf{18.42} \\
\midrule
RAG & Linear & 0.18 & -0.18 & -0.03 & -49.98 & -14.11 \\
    & \fixedbest{SLERP} & \fixedbest{0.19} & \fixedbest{-0.16} & \fixedbest{-0.01} & \fixedbest{-45.07} & \fixedbest{-5.69} \\
    & TIES   & 0.18 & -0.17 & -0.02 & -48.22 & -11.09 \\
    & \textbf{\simmerge{}} & \textbf{0.20} & \textbf{-0.15} & \textbf{-0.00} & \textbf{-42.01} & \textbf{-0.42} \\
\bottomrule
\end{tabular}
\label{tab:quad-summary}
\end{table}

Four-way merges (Table~\ref{tab:quad-summary}) further amplify the shortcomings of fixed operator choices. For several tasks, all fixed baselines incur large expert degradation and, in some cases, negative gains relative to auxiliaries, indicating harmful merges. Despite this increased complexity, \simmerge{} consistently remains the top-performing method across tasks, often matching or exceeding the best fixed operator while substantially reducing expert degradation. These results demonstrate that similarity-driven operator selection becomes increasingly important as the number of merged models grows.

\subsection{Per-Task Trends Across Merge Sizes}
\label{app:per-task-trends}

Tables~\ref{tab:pairwise-summary}--\ref{tab:quad-summary} report the exact per-task results for $k\in\{2,3,4\}$.
Here we complement those tables with per-task trend plots that visualize how performance evolves with merge size under two reference points:
(i) \emph{expert-relative} change ($\Delta_{\text{expert}}$), measuring preservation of task specialization, and
(ii) \emph{auxiliary-relative} change ($\Delta_{\text{aux}}$), measuring retention of useful off-domain capability.

\begin{figure}[h!]
  \centering
  \includegraphics[width=0.48\linewidth]{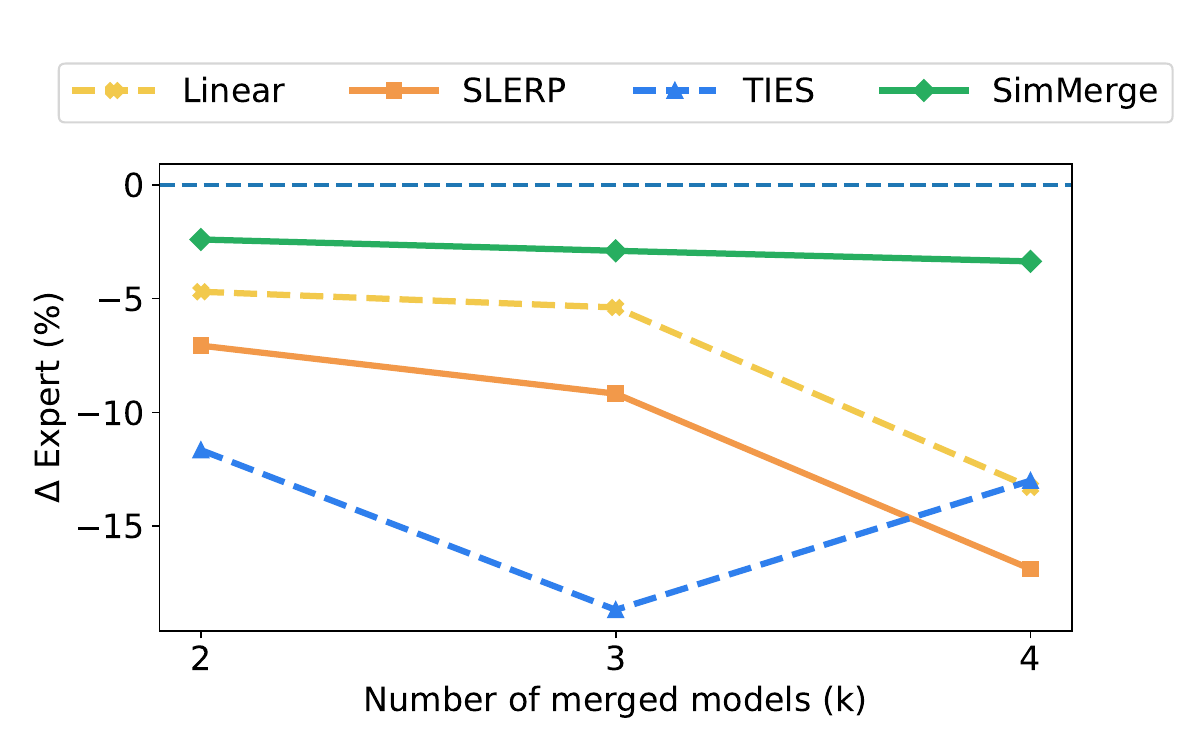}\hfill
  \includegraphics[width=0.48\linewidth]{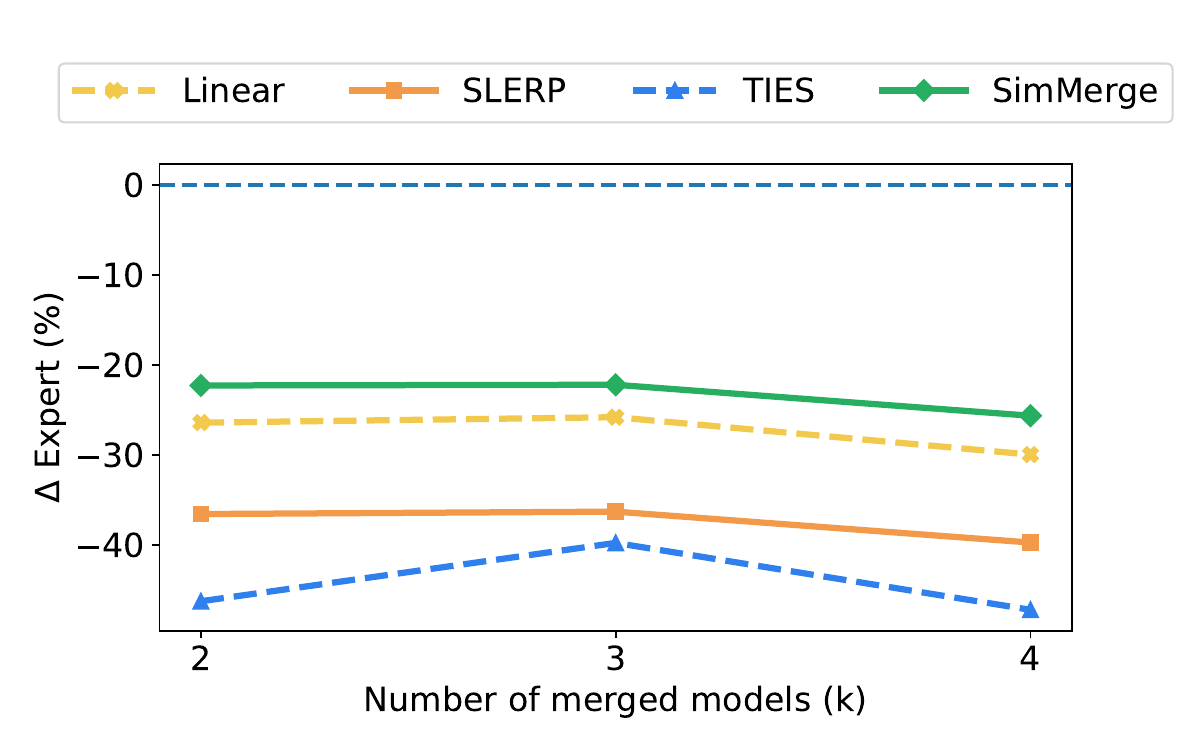}
  \caption{\textbf{Expert-relative trends.} Percentage change relative to the task expert ($\Delta_{\text{expert}}$) as the number of merged models increases. Left: Code. Right: Multilingual. The effect of merge size can be non-monotonic (often a drop at $k=3$ followed by partial recovery at $k=4$), reflecting higher-order interactions between fine-tuned updates. Across both domains, \simmerge{} remains closest to the expert across merge sizes.}
  \label{fig:app-deltaexp-code-multi}
\end{figure}

\textbf{Expert-relative trends (Code and Multilingual).}
Figure~\ref{fig:app-deltaexp-code-multi} shows $\Delta_{\text{expert}}$ as a function of $k$ for Code and Multilingual.
Across both domains, \simmerge{} remains closest to the expert at every merge size, forming the upper envelope among all methods.
Notably, the effect of increasing $k$ is not strictly monotonic: several fixed operators exhibit a pronounced degradation from $k=2$ to $k=3$ followed by partial recovery at $k=4$.
This non-monotonicity is consistent with multi-way composition dynamics, where the third model can introduce the first strong conflict between specialized updates, while adding a fourth model can partially cancel harmful directions under equal-weight merging.
The Multilingual domain is particularly sensitive: fixed operators separate more dramatically as $k$ increases, while \simmerge{} remains consistently closer to the expert.

\begin{figure}[h!]
  \centering
  \includegraphics[width=0.48\linewidth]{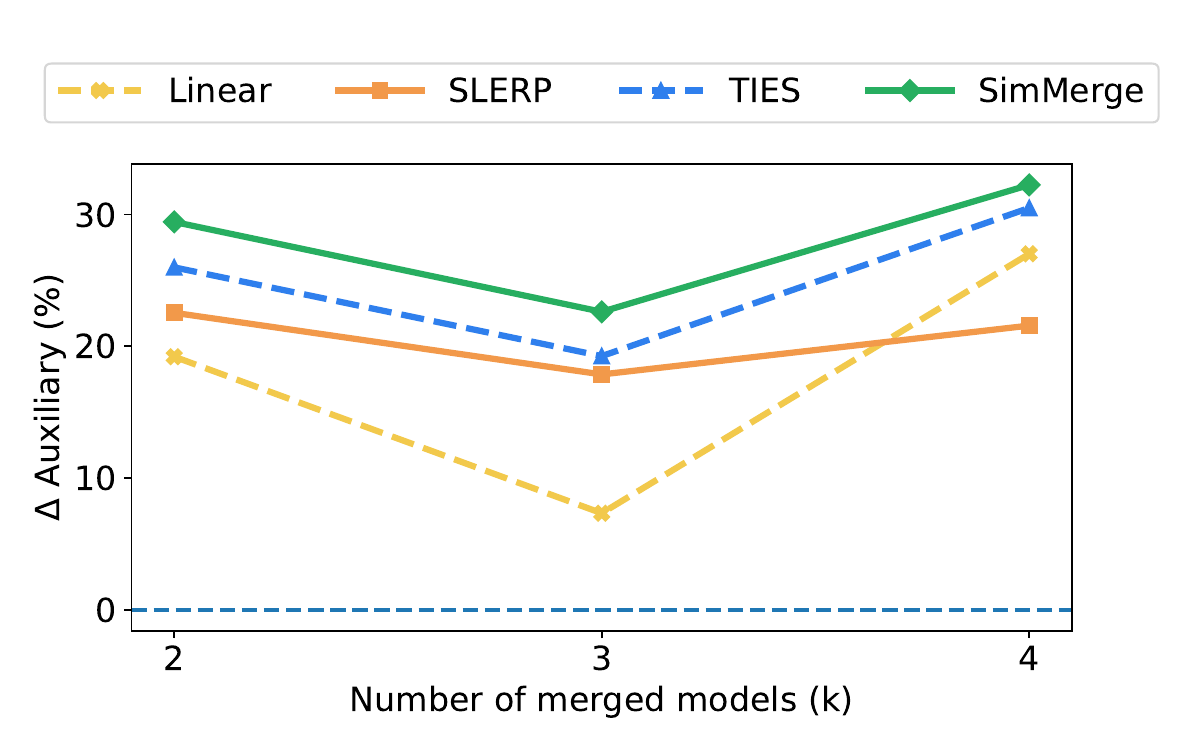}\hfill
  \includegraphics[width=0.48\linewidth]{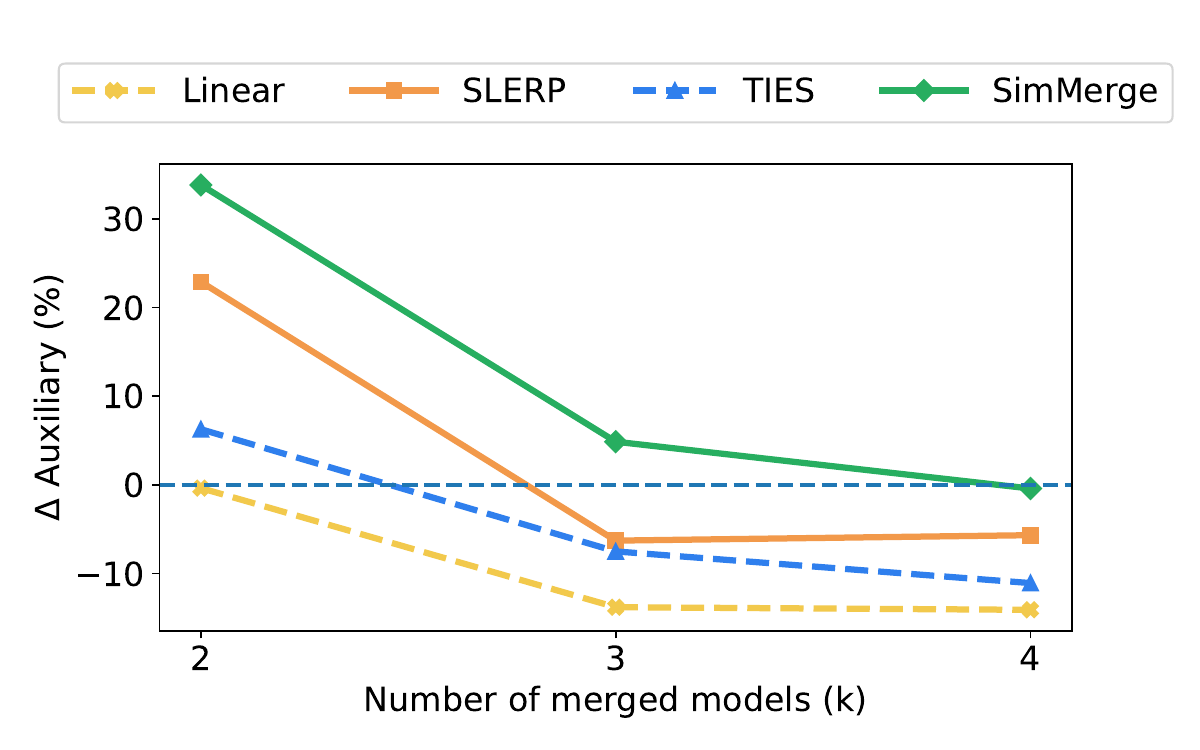}
  \caption{\textbf{Auxiliary-relative trends.} Percentage change relative to auxiliary baselines ($\Delta_{\text{aux}}$) as the number of merged models increases. Left: Math. Right: RAG. Fixed operators can exhibit diminishing or negative auxiliary gains at larger $k$, particularly in RAG. \simmerge{} more consistently preserves improvements over auxiliaries across merge sizes, indicating more robust retention of off-domain capability under multi-way composition.}
  \label{fig:app-deltaaux-math-rag}
\end{figure}

\textbf{Auxiliary-relative trends (Math and RAG).}
Figure~\ref{fig:app-deltaaux-math-rag} plots $\Delta_{\text{aux}}$ for Math and RAG.
As $k$ grows, auxiliary gains can shrink or even become negative for fixed operators, indicating that naive multi-way merges can underperform the auxiliary baseline.
This behavior is especially visible in RAG, where interference is strong and fixed operators often yield weakly positive or negative auxiliary percentage change.
In contrast, \simmerge{} more reliably maintains positive (or near-zero) auxiliary gains across merge sizes, suggesting better retention of off-domain capability while still limiting expert degradation (Figure~\ref{fig:app-deltaexp-code-multi}).

%\begin{figure}[t]
%  \centering
%  \includegraphics[width=0.9\linewidth]{figures/pairwise/rel_perf_vs_expert_per_task_and_overall.pdf}
%  \caption{%
%  Pairwise merges: percentage change in performance relative to the task expert, per task and averaged across tasks (Overall).%
%  Negative values indicate degradation relative to the expert.%
%  \simmerge{} consistently reduces expert degradation compared to the strongest fixed operator.}
%  \label{fig:pairwise-rel-expert}
%\end{figure}

%\begin{figure}[t]
%  \centering
%  \includegraphics[width=0.9\linewidth]{figures/pairwise/rel_perf_vs_aux_per_task_and_overall.pdf}
%  \caption{%
%  Pairwise merges: percentage change in performance relative to the auxiliary model(s), per task and averaged across tasks (Overall).%
%  Positive values indicate improvement over the auxiliary baseline.%
%  \simmerge{} achieves the largest gains over auxiliary models across tasks and in the overall average.}
%  \label{fig:pairwise-rel-aux}
%\end{figure}

\subsection{Overall Gap Closed summary}
\label{app:overall-gapclosed}

Figure~\ref{fig:pairwise-gap-closed} reports domain-level averages of \textit{Gap Closed}. To summarize overall performance with a single scalar, we take an unweighted macro-average across the four domain means on Code, Math, Multilingual, RAG.

\begin{table}[h!]
\caption{Domain-averaged \textsc{GapClosed} from Figure~\ref{fig:pairwise-gap-closed} and macro-average across domains.}
\centering
\small
\begin{tabular}{lccccc}
\toprule
Method & Code & Math & Multilingual & RAG & Macro avg. \\
\midrule
\linear{} & 69.0 & 61.5 & 37.1 & -0.4 & \fixedbest{41.8} \\
\slerp{}  & 53.3 & 72.2 & 12.9 & 23.0 & 40.4 \\
\ties{}   & 23.6 & 83.3 & -10.2 & 6.4  & 25.8 \\
\simmerge{} & 84.2 & 94.3 & 46.9 & 34.8 & \textbf{65.0} \\
\bottomrule
\end{tabular}
\label{tab:gapclosed-macro}
\end{table}

\begin{figure}[h!]
    \centering
    \includegraphics[width=0.9\linewidth]{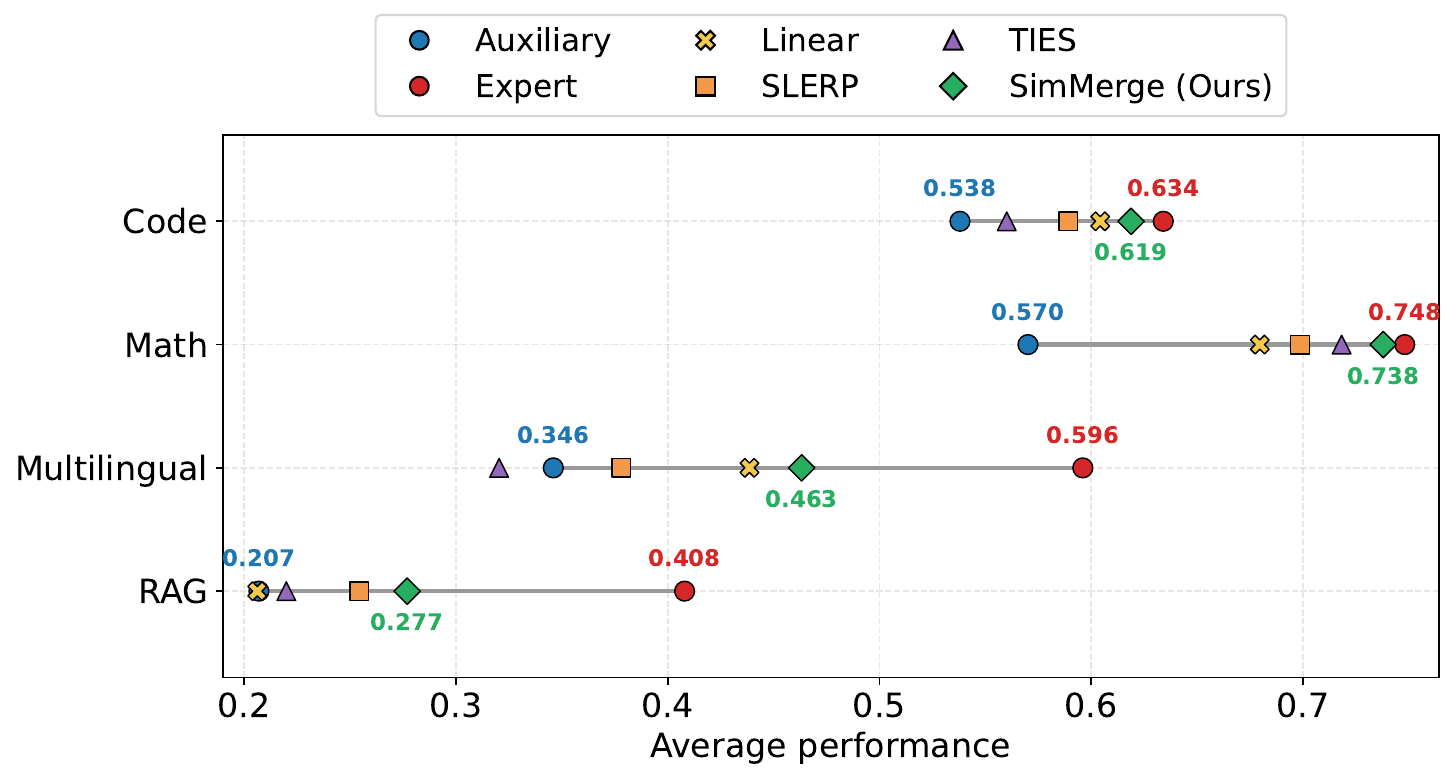}
    \caption{Absolute average performance per domain for auxiliary, expert, and merged models. This anchors the normalized metrics to the original task scales.}
    \label{fig:absolute-dumbbell}
\end{figure}

To complement Table~\ref{tab:gapclosed-macro} and Figure~\ref{fig:pairwise-gap-closed}, Figure~\ref{fig:absolute-dumbbell} reports the corresponding average performance for the auxiliary, expert and merged models.

Figure~\ref{fig:absolute-dumbbell} makes two points explicit on the original task scales.
First, the expert-auxiliary gap differs substantially by domain. For instance, the expert improves over the auxiliary from $0.538$ to $0.634$ on Code, from $0.570$ to $0.748$ on Math, from $0.346$ to $0.596$ on Multilingual, and from $0.207$ to $0.408$ on RAG.
Second, \simmerge{} consistently produces merged models that move toward the expert while improving over the auxiliary baseline across all four domains.
In the pairwise setting, \simmerge{} achieves the highest mean performance in every domain, reaching $0.62$ on Code, $0.74$ on Math, $0.46$ on Multilingual, and $0.28$ on RAG, outperforming the best fixed operator in each case.
By contrast, the strongest fixed operator is domain-dependent: Linear on Code and Multilingual, TIES on Math, and SLERP on RAG, reinforcing that no single merge rule dominates across tasks.
This absolute view also clarifies the trade-off: \simmerge{} improves off-domain performance while incurring smaller degradation relative to the expert than fixed baselines.

\subsection{111B task-level results for 3-way merges}
\label{app:111b}

\begin{figure}[h!]
  \centering
  \includegraphics[width=0.90\linewidth]{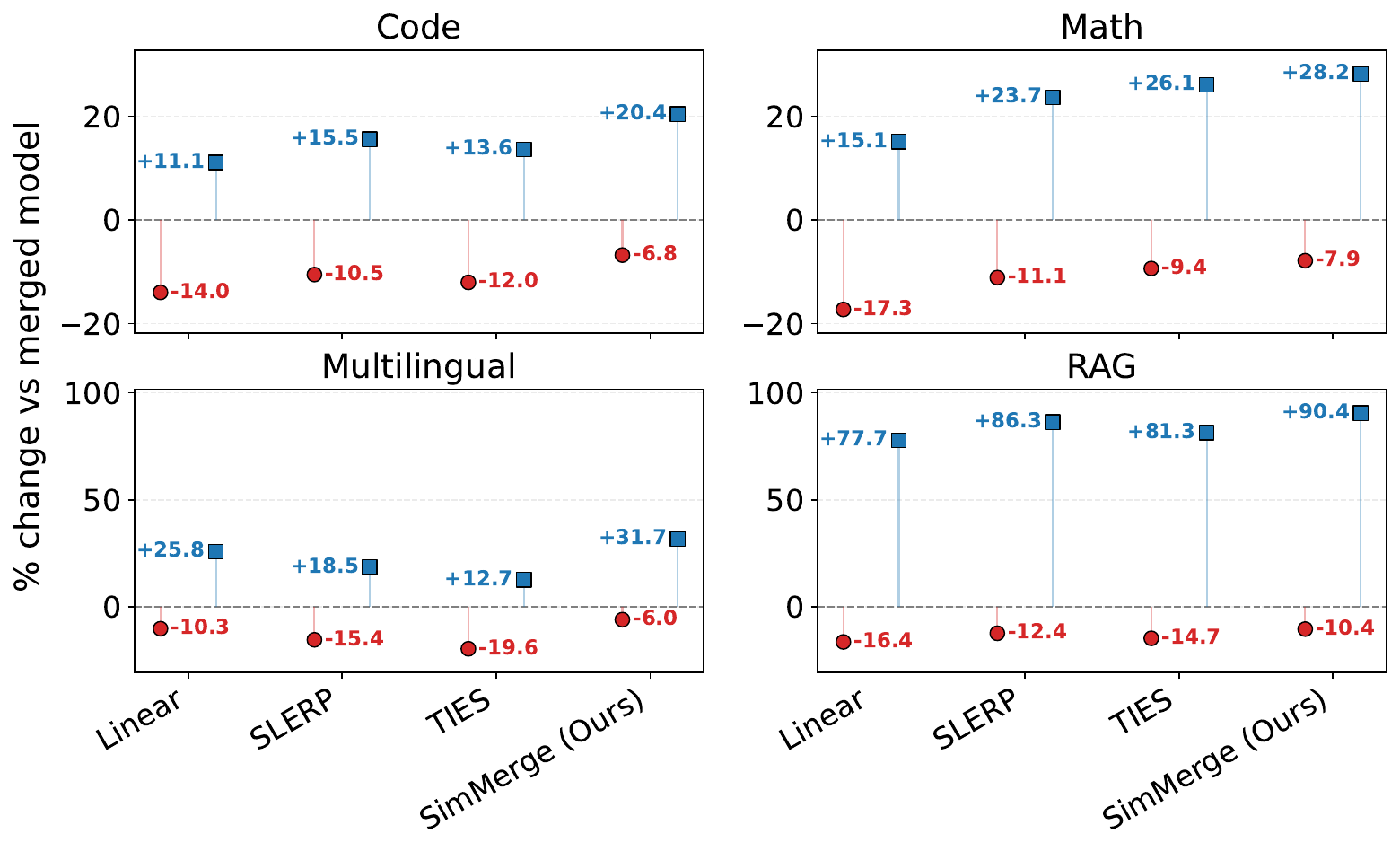}
  \caption{Per-task percentage change in performance for each merge method for 3-way merges at 111B. Blue markers show $\Delta_{\text{aux}}$ (change vs.\ auxiliary; higher is better) and red markers show $\Delta_{\text{expert}}$ (change vs.\ task expert; closer to $0$ indicates less degradation), as defined in Section~\ref{sec:exp-metrics}.}
  \label{fig:111b-percent-by-domain}
\end{figure}

Figure~\ref{fig:111b-percent-by-domain} breaks down 3-way merging results at 111B by domain, reporting both auxiliary-relative gains $\Delta_{\text{aux}}$ and expert-relative degradation $\Delta_{\text{expert}}$ as defined in Section~\ref{sec:exp-metrics}.
%The plotted quantities correspond to $\Delta_{\text{aux}}$ in blue and $\Delta_{\text{expert}}$ in red.
Across all four domains, \simmerge{} achieves the strongest expert-auxiliary trade-off. It produces the smallest degradation relative to the expert in every domain, and it also delivers the largest gain over auxiliaries.
For example, on Code it reduces expert-relative degradation to $-6.8\%$ while improving over auxiliaries by $+20.4\%$.
On RAG, it achieves a large auxiliary gain of $+90.4\%$ while keeping expert degradation at $-10.4\%$.

\begin{figure}[h!]
  \centering
  \includegraphics[width=0.90\linewidth]{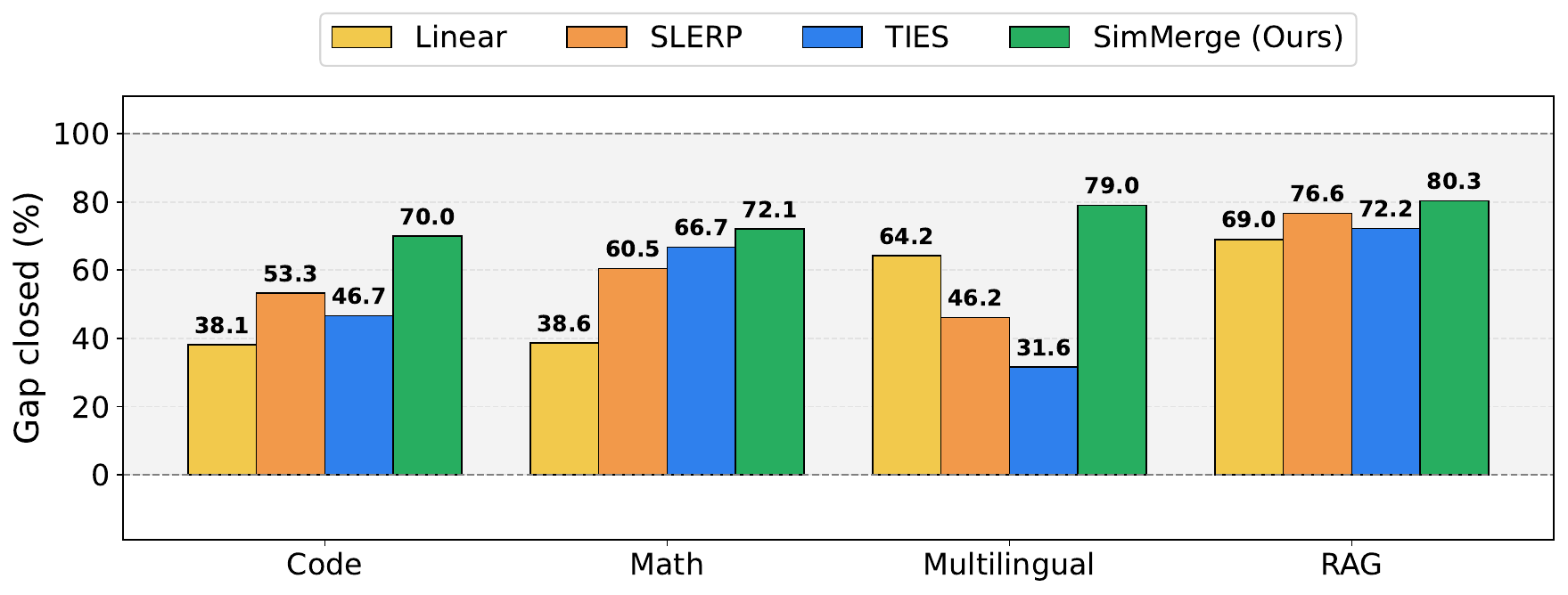}
  \caption{\textsc{GapClosed} for 3-way merges at 111B across Code, Math, Multilingual, and RAG.}
  \label{fig:111b-gap-closed}
\end{figure}

Figure~\ref{fig:111b-gap-closed} reports the same comparison using \textsc{GapClosed}, which normalizes performance so that 0\% corresponds to the auxiliary baseline and 100\% corresponds to the expert. \simmerge{} achieves the highest \textsc{GapClosed} in every domain.
The best fixed operator varies by domain, but \simmerge{} remains best overall, reaching 69.0 on Code, 76.6 on Math, 70.0 on Multilingual, and 80.3 on RAG.

\section{Tail Effects and Similarity Correlations}
\label{sec:appendix-tails}

\begin{figure}[h!]
    \centering
    \begin{subfigure}[t]{0.95\linewidth}
        \centering
        \includegraphics[width=\linewidth]{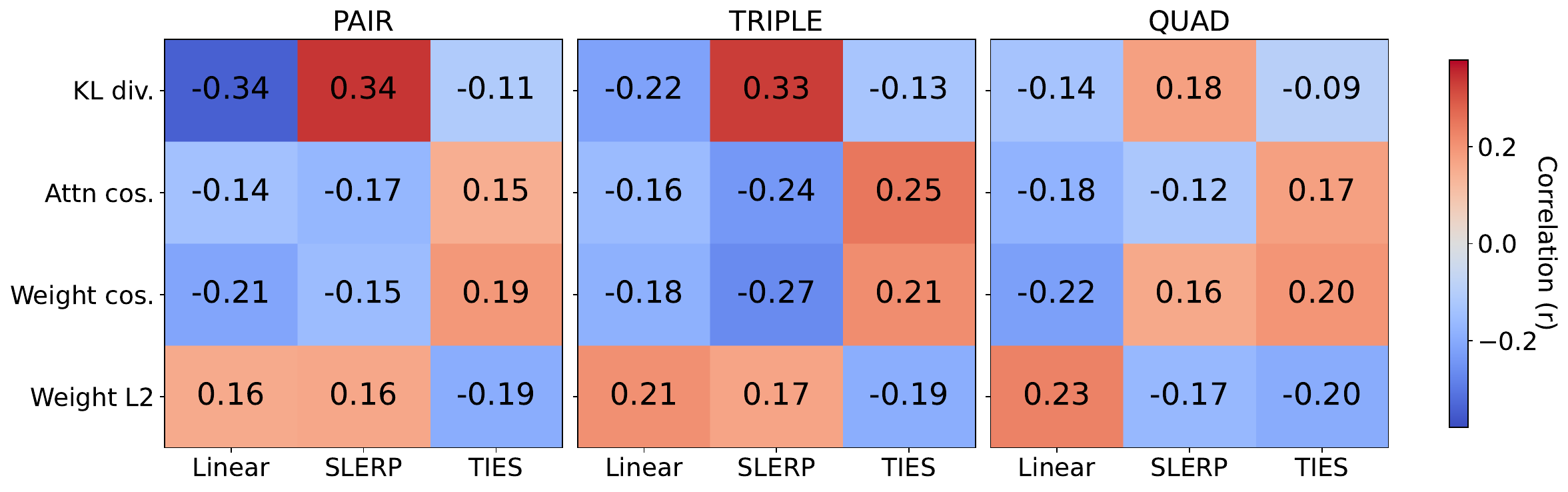}
        \caption{Spearman}
        \label{fig:corr-spearman}
    \end{subfigure}
    \hfill
    \begin{subfigure}[t]{0.95\linewidth}
        \centering
        \includegraphics[width=\linewidth]{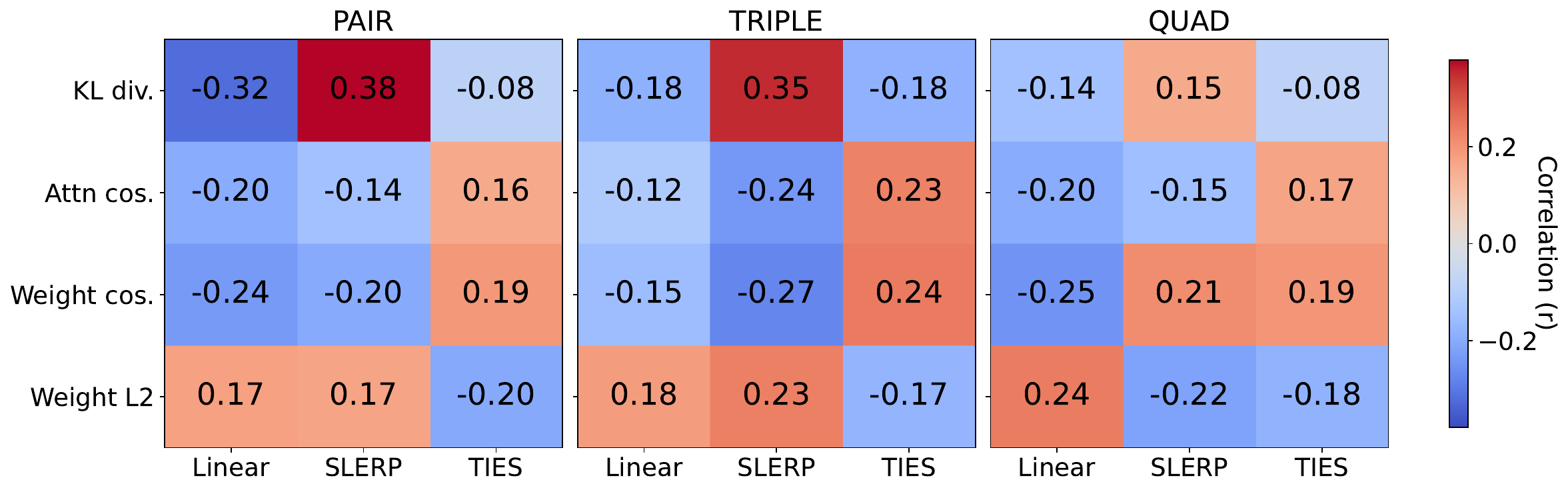}
        \caption{Pearson}
        \label{fig:corr-pearson}
    \end{subfigure}
    \caption{Correlation between similarity signals and merge performance for \linear{}, \slerp{} and \ties{} across pairwise, triple, and quadruple merges. Spearman and Pearson correlations exhibit consistent patterns, indicating robust, operator-specific similarity regimes.}
    \label{fig:corr-heatmaps}
\end{figure}

We begin by examining how similarity signals correlate with merge operator performance.
Figure~\ref{fig:corr-heatmaps} shows Pearson and Spearman correlations between similarity features and performance outcomes for \linear{}, \slerp{} and \ties{} across pairwise (PAIR), triple (TRIPLE), and quadruple (QUAD) merges.

Several consistent patterns emerge. Across all merge settings, different similarity features are predictive of success for different
operators, often with opposing signs.
KL divergence is positively correlated with \slerp{} performance but negatively correlated with \linear, while weight cosine similarity exhibits the opposite pattern. Attention-based cosine similarity shows positive correlation with \ties, whereas weight $\ell_2$ distance is most predictive of \linear's success.

The close agreement between Pearson and Spearman correlations indicates that these relationships are robust and largely monotonic rather than being driven by a small number of outliers. Importantly, no single similarity feature correlates positively with all operators, suggesting that merge quality is inherently regime-dependent.

\begin{figure*}[h!]
    \centering
    \begin{subfigure}[t]{0.49\textwidth}
        \includegraphics[width=\linewidth]{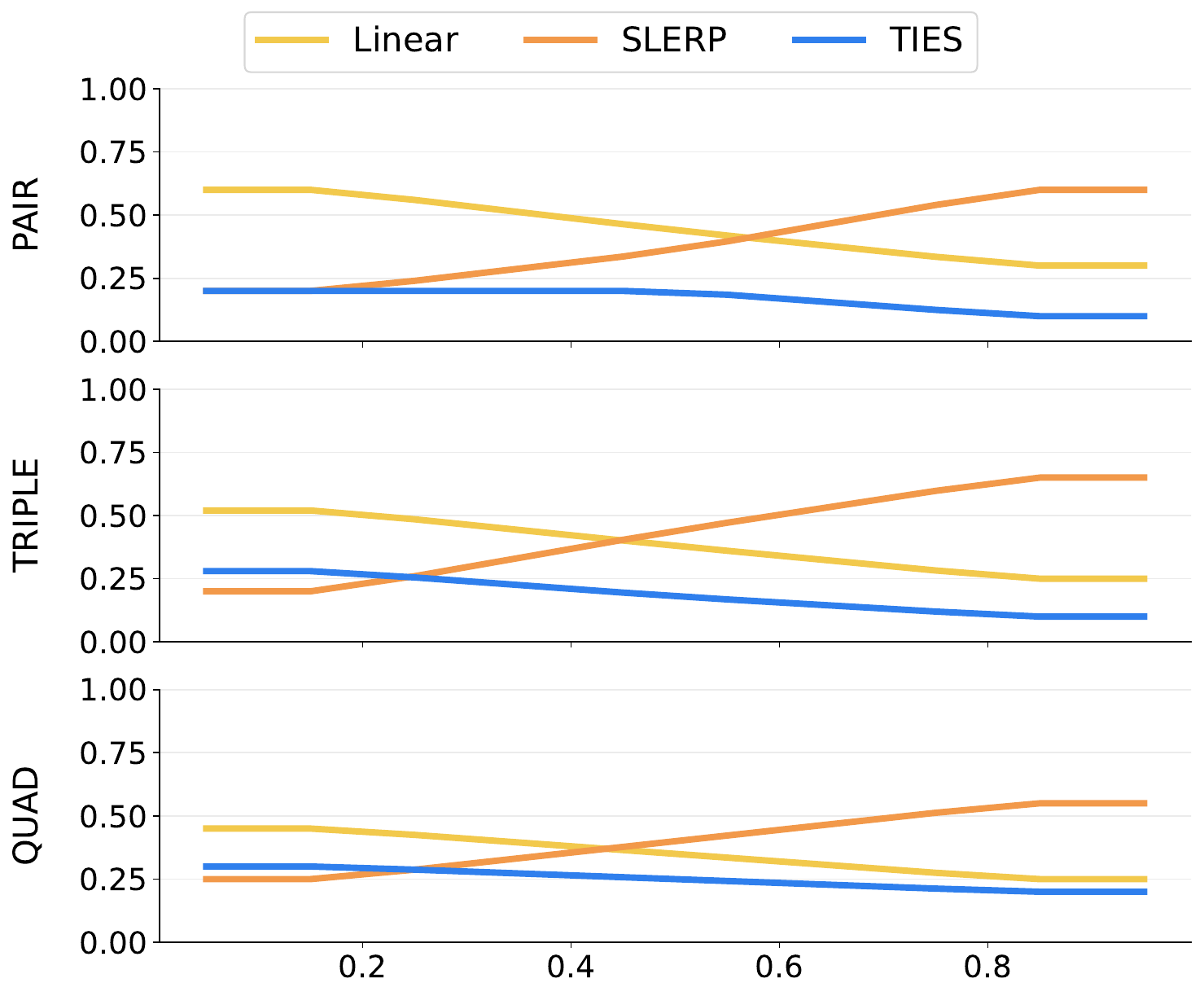}
        \caption{KL divergence (logits)}
        \label{fig:pwin_kl}
    \end{subfigure}
    \hfill
    \begin{subfigure}[t]{0.49\textwidth}
        \includegraphics[width=\linewidth]{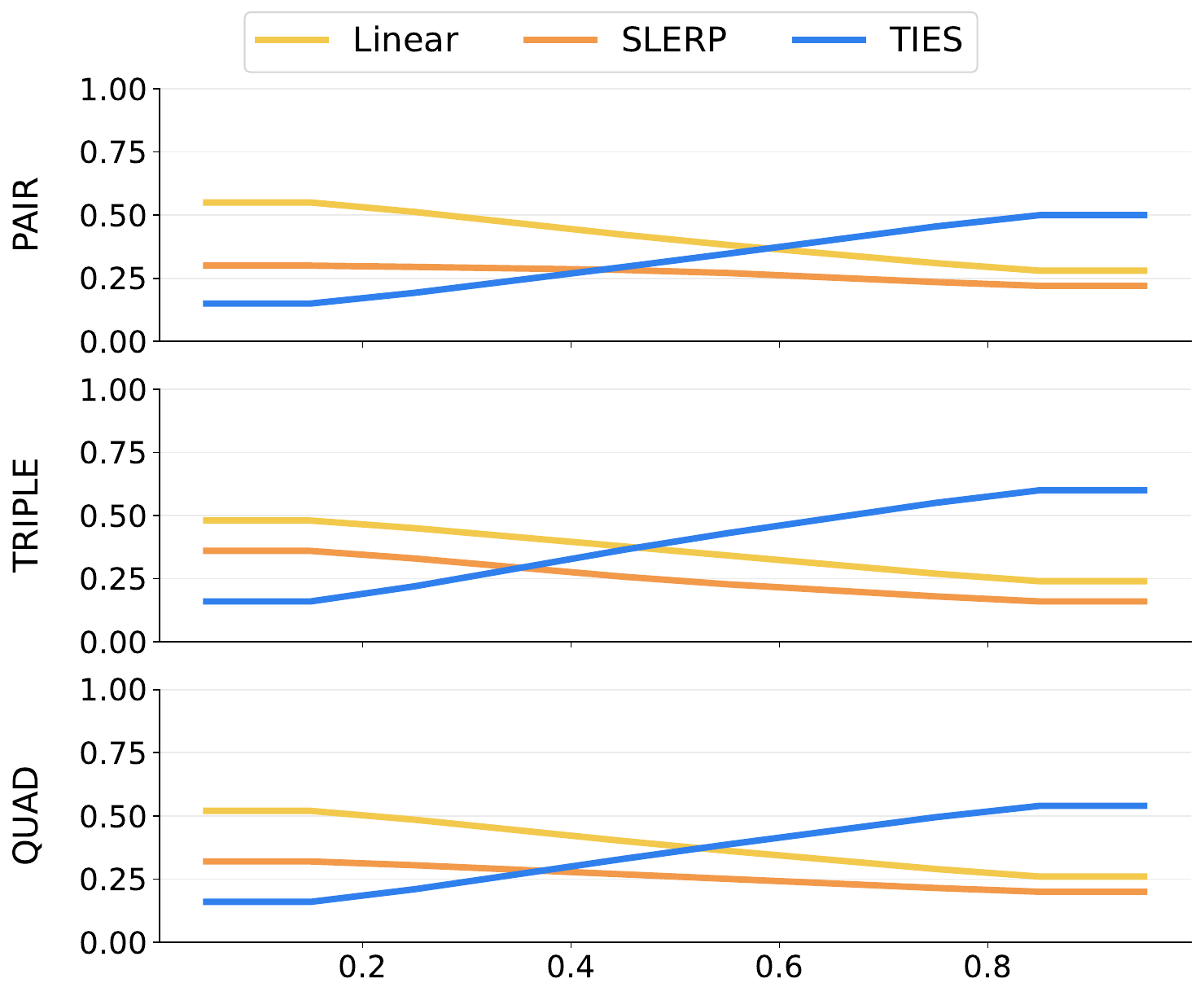}
        \caption{Attention cosine similarity}
        \label{fig:pwin_attncos}
    \end{subfigure}

    \vspace{0.6em}

    \begin{subfigure}[t]{0.49\textwidth}
        \includegraphics[width=\linewidth]{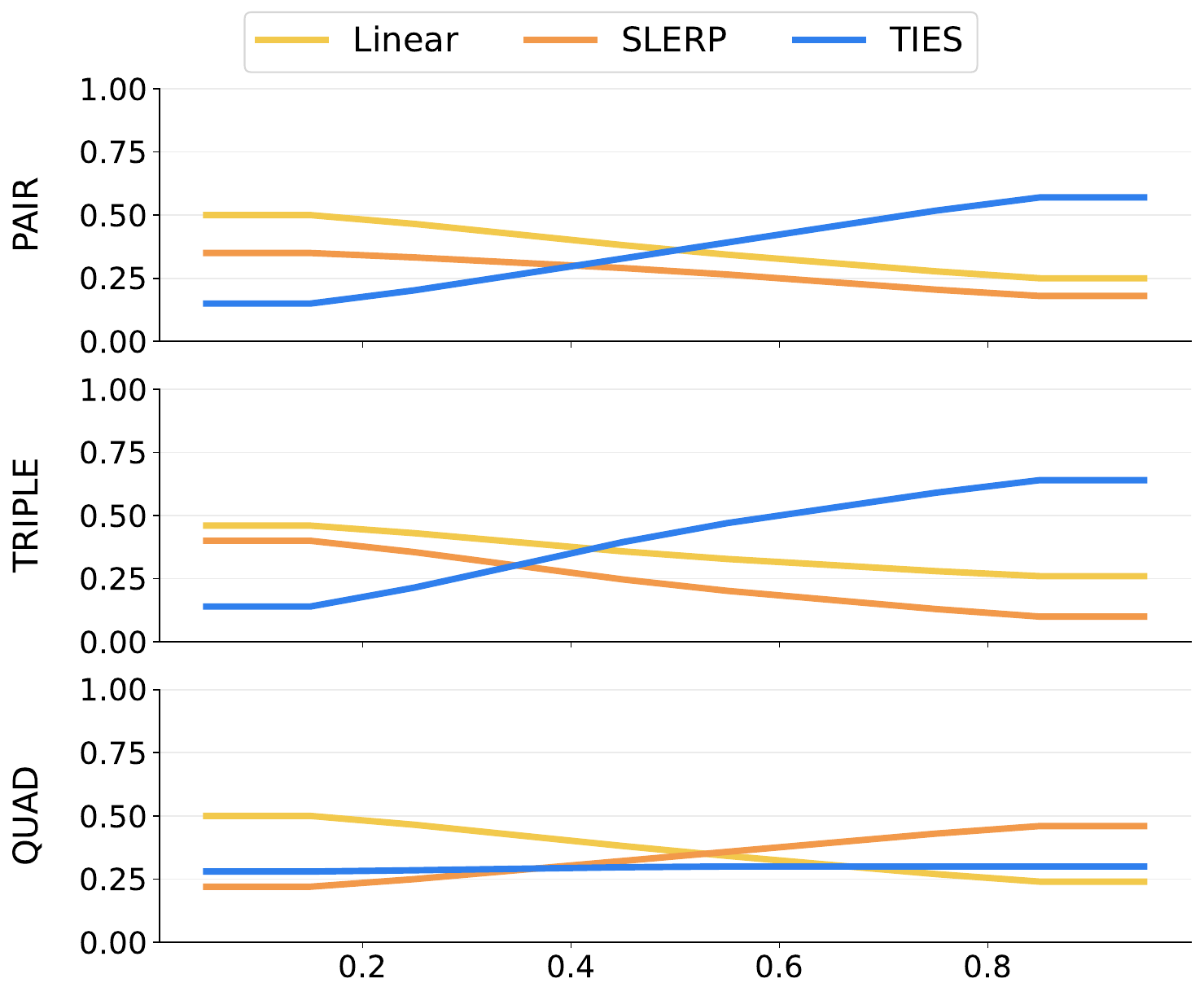}
        \caption{Weight cosine similarity}
        \label{fig:pwin_wcos}
    \end{subfigure}
    \hfill
    \begin{subfigure}[t]{0.49\textwidth}
        \includegraphics[width=\linewidth]{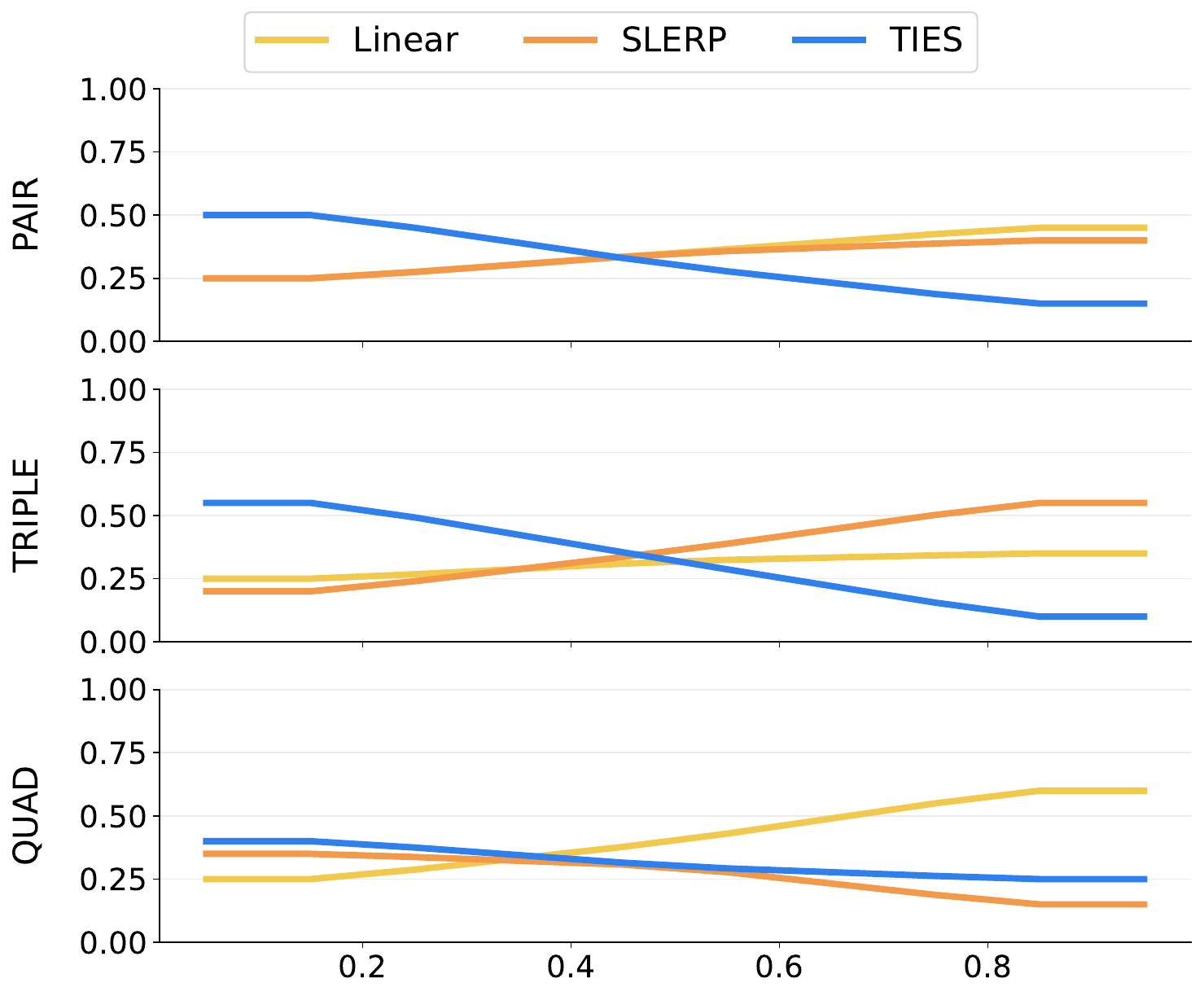}
        \caption{Weight $\ell_2$ distance}
        \label{fig:pwin_l2}
    \end{subfigure}

    \caption{\textbf{Percentile-bin win trends.} For each case (PAIR/TRIPLE/QUAD), we map each probe metric to its empirical percentile and compute $P(\text{win})$ for each merge operator within equal-mass percentile bins. This makes trends comparable across merges and across metrics with different raw scales.}
    \label{fig:pwin_percentile_trends}
\end{figure*}

\textbf{Trends in percentile-bin.}
The percentile-bin curves in Figure~\ref{fig:pwin_percentile_trends} visualize how each merge operator's win probability ($P(\mathrm{win})$) varies as a function within-case percentiles. Overall, the directions of these trends are consistent with the winner-metric correlations shown in Figure~\ref{fig:corr-heatmaps}.

Figure~\ref{fig:pwin_kl} shows that moving from low to high KL percentiles increases $P(\mathrm{win}=\slerp{})$ while decreasing $P(\mathrm{win}=\linear{})$ while $\ties{}$ is weakly decreasing or near-flat.

In Fig.~\ref{fig:pwin_attncos}, $P(\mathrm{win}=\ties{})$ increases monotonically with attention cosine similarity percentiles, while $P(\mathrm{win}=\linear{})$ decreases, $\slerp{}$ tends to decrease, most clearly in the triple merge setting.

Figure~\ref{fig:pwin_wcos} shows that higher weight-cosine percentiles favor $\ties{}$ and disfavor $\linear{}$ across merge settings. Notably, $\slerp{}$ decreases with weight cosine in pairwise and triple merges but increases in QUAD, consistent with the corresponding sign flip observed in Figure~\ref{fig:corr-heatmaps}.
This pattern is consistent with spherical interpolation becoming more reliable when strong mutual parameter alignment is present in four-way merges.

Figure~\ref{fig:pwin_l2} shows that increasing weight $\ell_2$ percentiles increase $P(\mathrm{win}=\linear{})$ and decrease $P(\mathrm{win}=\ties{})$ across merge settings. $\slerp{}$ increases with $\ell_2$ distance in PAIR and TRIPLE but decreases in QUAD, again mirroring the sign changes in Figure~\ref{fig:corr-heatmaps}.
This pattern is consistent with a geometric interpretation of the merge operators. Large weight $\ell_2$ distance reflects substantial parameter magnitude mismatch between models.
In such regimes, Linear interpolation, which does not rely on directional alignment, tends to be more robust, while TIES degrades due to increased trimming under magnitude differences.
SLERP improves with increasing $\ell_2$ distance in pairwise and triple merges but degrades in the quadruple setting, where averaging across multiple directions becomes less stable.

\textbf{Effects of merge size.}
Comparing the panels within each subfigure in Figure~\ref{fig:pwin_percentile_trends}, QUAD trends are often flatter, indicating weaker dependence on similarity percentiles. Additionally, $\textsc{SLERP}$ exhibits two notable merge-size-dependent reversals: (i) a positive association with weight cosine similarity in QUAD (Figure~\ref{fig:pwin_wcos}), and (ii) a negative association with weight $\ell_2$ distance in QUAD (Figure~\ref{fig:pwin_l2}).
These effects indicate that multi-model geometry introduces interactions beyond those captured by pairwise relationships.

\textbf{Tail effects and operator robustness.}
While similarity-conditioned trends describe average behavior across similarity regimes, they do not capture how performance is distributed across individual merge instances. In particular, an operator may perform well on average while still failing catastrophically on a nontrivial fraction of cases. To characterize this behavior, we analyze \emph{tail effects}, which quantify whether a method's wins are concentrated in favorable regimes or whether it frequently appears among the worst-performing
outcomes.

For each merge method, metric, and merge setting, we define the tail effect as
\begin{equation}
\Delta P(\mathrm{win}) = P(\text{top }20\%) - P(\text{bottom }20\%)
\label{eq:tail_effect}
\end{equation}
where $P(\text{top }20\%)$ denotes the probability that the method ranks in the top quintile of outcomes, and $P(\text{bottom }20\%)$ denotes the probability of ranking in the bottom quintile.
A large positive value indicates that a method consistently wins in favorable regimes while rarely failing badly, whereas values near zero or negative indicate brittle behavior with frequent severe failures.

\begin{figure}[h!]
    \centering
    \includegraphics[width=0.9\linewidth]{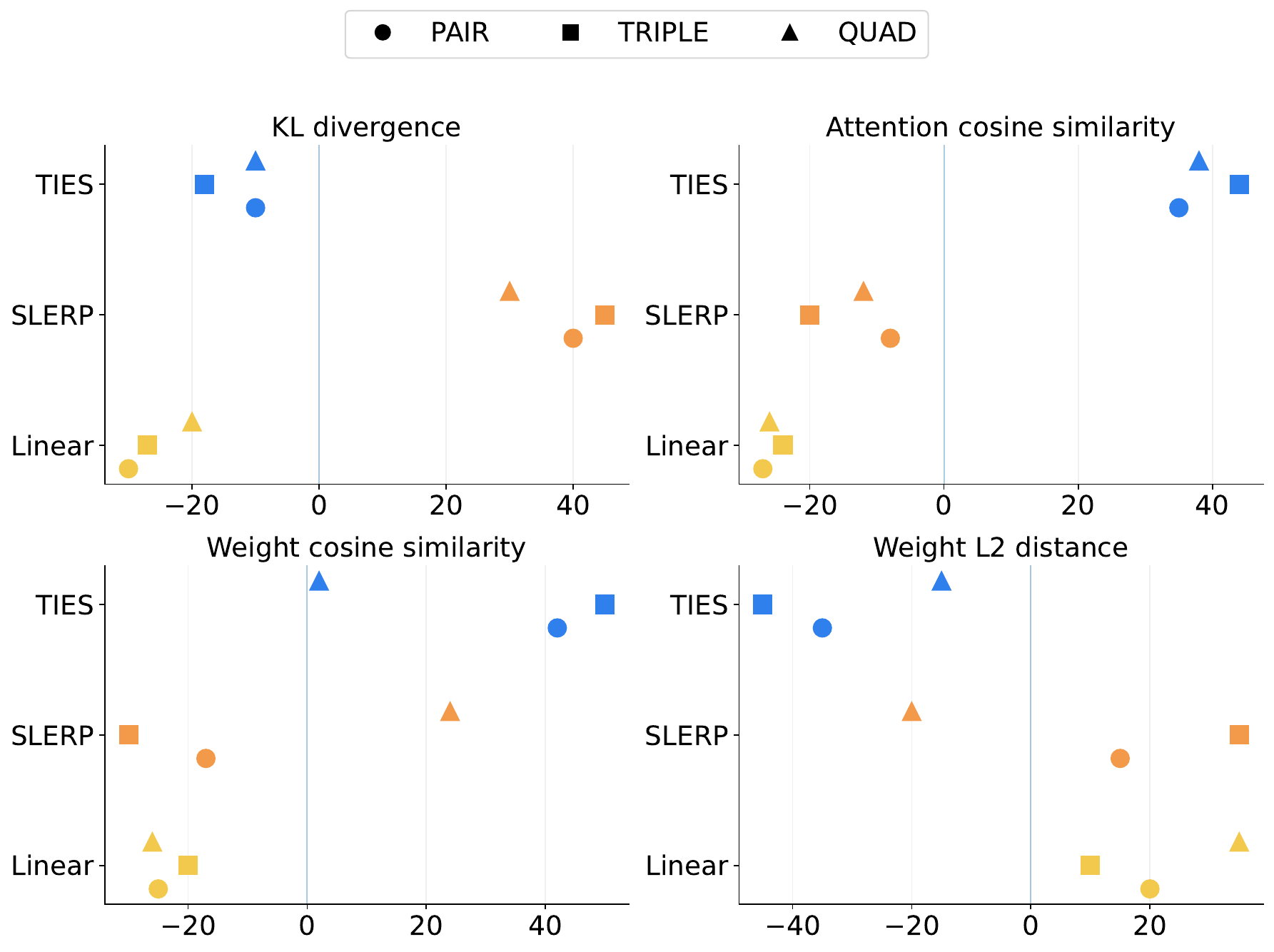}
    \caption{Tail effects of merge operators across similarity metrics and merge settings. Each point shows the tail-effect score (Eq.~\ref{eq:tail_effect}). Markers indicate pairwise, triple, and quad merges.}
    \label{fig:tail-effect-dot}
\end{figure}

Figure~\ref{fig:tail-effect-dot} visualizes tail effects across similarity metrics and merge settings.
Fixed operators exhibit strong and highly metric-dependent tail behavior.
For example, \linear{} shows positive tail effects in regimes characterized by small weight distances, but negative or near-zero tail effects under KL divergence and attention-based similarity.
Conversely, \ties{} concentrates wins when weight cosine or attention similarity is high, but frequently occupies the bottom tail outside these regimes. \slerp{} exhibits mixed behavior, with tail effects that change sign depending on both the similarity metric and the merge setting.

As merge complexity increases from pairwise to quad settings, tail effects generally become more pronounced. This indicates that applying a single operator uniformly across increasingly heterogeneous collections
of models amplifies the risk of severe failures, even when average performance remains competitive. These tail failures explain why fixed operators can appear strong under aggregate metrics yet behave unreliably in practice.

Taken together, similarity-conditioned trends and tail effects show that merge operator effectiveness is inherently regime-dependent.
Each operator succeeds only within specific similarity regimes and exhibits sharp failures outside them, leading to brittle behavior when a single rule is applied universally.
By identifying these regimes through similarity signals and selecting operators on a per-instance basis, \simmerge{} avoids unfavorable tails and achieves robust merging behavior across tasks and merge settings.

\end{document}